\documentclass[runningheads]{llncs}
\usepackage{scalerel,graphicx,amssymb,amsmath,mathtools,wasysym,floatrow}
\usepackage[outercaption]{sidecap} 
\usepackage[ruled,vlined,linesnumbered]{algorithm2e}
\usepackage[dvipsnames]{xcolor}
\usepackage[colorinlistoftodos]{todonotes}
\usepackage{txfonts}
\usepackage{tikz}
\usepackage{caption}
\usepackage{subcaption}
\usepackage{wrapfig,lipsum}
\usepackage{floatrow}
\usetikzlibrary{arrows}

\DeclareMathOperator*{\argmax}{arg\,max}

\DeclareMathSymbol{\mlq}{\mathord}{operators}{``}
\DeclareMathSymbol{\mrq}{\mathord}{operators}{`'}
\newcommand\Usf{\mathcal{U}_{\sitf}}
\newcommand\Ul{\mathcal{L}}

\newcommand\prfx{\mathit{prfx}}

\newcommand\val{\mathit{values}}

\newcommand\eq{\mathcal{EQ}}
\newcommand\SF{\boldsymbol{SF}}
\newcommand\sfunc{ \# }
\newcommand\V{\boldsymbol{V}}
\newcommand\U{\boldsymbol{U}}

\newcommand\Uinst{\mathcal{U}_{instance}}

\newcommand\sitf{\mathit{sf}}

\newcommand\csf{\mathit{csf}}

\newcommand{\car}{\ensuremath{%
		\bullet\kern-5pt
		\raise1pt\hbox{$\mathord{\rightarrow}$}}}
\def\cra{\hbox{$\multimapdotinv$}\kern-7.5pt\hbox{$\rightarrow$}}

\newif\ifComments
\Commentstrue

\begin{document}
	\title{Feature Recommendation for 
	Structural Equation Model Discovery in Process Mining}
	\author{Mahnaz Sadat Qafari \and
		Wil van der Aalst}
	\authorrunning{M. S. Qafari et al.}
	%
	\institute{Rheinisch-Westfälische Technische Hochschule Aachen(RWTH), Aachen, Germany \\
		\email{m.s.qafari@pads.rwth-aachen.de,wvdaalst@pads.rwth-aachen.de}}
	
	\maketitle 
	\begin{abstract}
	Process mining techniques can help organizations to improve the operational processes. Organizations can benefit from process mining techniques in finding and amending the root causes of performance or compliance problems. Considering the volume of the data and the number of features captured by the information system of today's companies, the task of discovering the set of features that should be considered in root cause analysis can be quite involving. In this paper, we propose a method for finding the set of (aggregated) features with a possible effect on the problem. 
	The root cause analysis task is usually done by applying a machine learning technique to the data gathered from the information system supporting the processes. To prevent mixing up correlation and causation, which may happen because of interpreting the findings of machine learning techniques as causal, we propose a method for discovering the structural equation model of the process that can be used for root cause analysis. We have implemented the proposed method as a plugin in ProM and we have evaluated it using two real and synthetic event logs. These experiments show the validity and effectiveness of the proposed methods.

\end{abstract}
	
\section{Introduction}
Organizations aim to improve operational processes to serve customers better and to become more profitable. To this goal, they can utilize process mining techniques in many steps, including identifying friction points in the process, finding the root causes of each friction point, estimating the possible impact of changing each factor on the process performance, and also planning process enhancement actions. Today, there are several robust techniques for process monitoring and finding their friction points, but little work on \textit{root cause analysis}. So, in this paper, we focus on root cause analysis and investigating the impact of interventions.


Processes are complicated entities involving many steps, where each step itself may include many influential factors and features. Moreover, not just the steps but also the order of the steps that are taken for each process instance may vary, which results in several process instance variants. This makes it quite hard to identify the set of features that influence a problem. In the literature related to root cause analysis in the field of process mining, it is usually assumed that the user provides the set of features that have a causal relationship with the observed problem in the process (see for example \cite{de2016general,qafari2021case}). To overcome this issue, we have proposed a mechanism that not only helps to identify the set of features that may have a causal relationship with the problem, but also the values of these features that are more prone to causing the problem.


Traditionally, the task of finding the root cause of a problem in a process is done in two steps; first gathering process data from the event log, and then applying data mining and machine learning techniques. Although the goal is to perform root cause analysis, a naive application of such techniques often leads to a mix-up of correlation and causation. It is easy to find correlations, but very hard to determine causation. Consequently, process enhancement based on the results of such approaches does not always lead to any process improvements. 


Consider the following three scenarios:
\begin{enumerate}
	\item [(i)] In an online shop, it has been observed that the possibility of delay in delivery is much higher if some specific resources are responsible for them.
	\item [(ii)] In a consultancy company, there are deviations in some cases and those cases have been done mainly by the employees who are most experienced.
	\item [(iii)] In an IT company, it has been observed that the higher the number of resources assigned to a task, the longer it takes.
\end{enumerate}
The following possibly incorrect conclusions can be made if these observed correlations are observed as causal relationships.
\begin{itemize}
    \item In the online shop scenario, the responsible resources are causing the delays.
    \item In the second scenario, we may conclude that over time the employees get more and more reckless, and consequently the rate of deviations increases.
    \item In the IT company, we may conclude that the more people working on a project, the more time is spent on team management and communication, which prolongs the project unnecessarily.
\end{itemize}
However, correlation does not mean causation. We can have a high correlation between two events when they are caused by a possibly unmeasured (hidden) common cause (set of common causes), which is called a \emph{confounder}. For example, in the first scenario, the delayed deliveries are mainly for the bigger size packages which are usually assigned to specific resources. Or, in the second scenario, the deviations happen in the most complicated cases that are usually handled by the most experienced employees. In the third scenario, maybe both the number of employees working on a project and the duration of a project are highly dependent on the complexity of the project. As it is obvious from these examples, changing the process based on the observed correlations not only leads to no improvement but also may aggravate the problem (or create new problems). 

Randomized experiments and the theory of causality are two general frameworks for finding the causes of a problem \cite{pearl2009causality,peters2017elements}. The randomized experiment provides the most robust and reliable method for making causal inferences and statistical estimates of the effect of an intervention. This method involves randomly setting the values of the features that have a causal effect on the observed problem and monitoring the effects. applying randomized experiments in the context of processes is usually too expensive (and sometimes unethical) or simply impossible. The other option for anticipating the effect of any intervention on the process is using a \emph{structural causal model} \cite{pearl2009causality,peters2017elements}. In this method, first, the causal mechanism of the process features is modeled by a conceptual model and then this model is used for studying the effect of changing the value of a process features.

\begin{figure}
	\includegraphics[width=120mm]{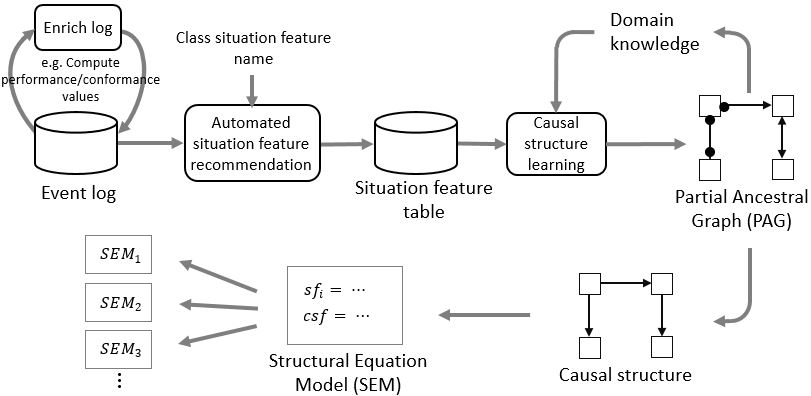}
	\caption{The general structural causal equation discovery.}\label{pic::general}
\end{figure}

This paper is an extension of~\cite{me2020}, where we have proposed a framework for root cause analysis using structural equation modeling. Here we address one of the main issues in this framework. Finding the features that may have a causal effect on the problem often requires substantial domain knowledge. Considering the variety of the feature values in a process, even when having extensive domain knowledge, it may not be easy to determine the values of the features that have the strongest effect on the problem. So, we propose a method for finding a set of features and feature value pairs that may contribute the most to the problem. Also, we add aggregated features to the features that can be extracted from the event log. This makes the method capable of analyzing more scenarios. The framework explained in this paper includes the following steps:
\begin{itemize}
    \item As a preprocessing step, the event log is enriched by several process-related features. These features are derived from different data sources like the event log, the process model, and the conformance checking results. Also, here we consider the possibility of adding aggregated features to the event log regarding the time window provided by the user.
    \item A set of pairs of the form (feature, feature value) are recommended to the user. Such pair include the features that might have a causal relationship with the problem and those values of them that possibly contribute more to the problem. Users can modify this set of features that have been identified automatically or simply ignore it and provide another set of features.
    \item The next step is creating a target-dependent data table, which we call it \emph{situation feature table}.
    \item This step involves generating a graphical object encoding the structure of causal relationships among the process features. This graphical object can be provided by the customer or be inferred from the observational data using a causal structure learning algorithm, also called \emph{search algorithm}. The user can modify the resulting graphical object by adding domain knowledge as an input to the search algorithm or by modifying the discovered graph.
    \item The last step involves estimating the strength of each discovered causal relationship and the effect of an intervention on any of the process features on the identified problem.
\end{itemize}  

In Figure \ref{pic::general}, the general overview of the proposed approach is presented.

The remainder of the paper is organized as follows. In Section \ref{sec::ex}, we start with an example. We use this example as the running example throughout this paper. In Section~\ref{rw}, we present some of the related work. The corresponding process mining and causal inference theory preliminaries are presented in Section \ref{prel} and, in Section \ref{app}, an overview of the proposed approaches for feature recommendation and causal equation model discovery is presented. In Section ~\ref{sec::er}, the assumptions and the design choices in the implemented plugin and the experimental results of applying it on synthetic and real event logs are presented. Finally, in Section~\ref{sec::conclution}, we summarize our approach and its applications.

\section{Motivating Example}\label{sec::ex}
\begin{figure}
	\includegraphics[width=110mm]{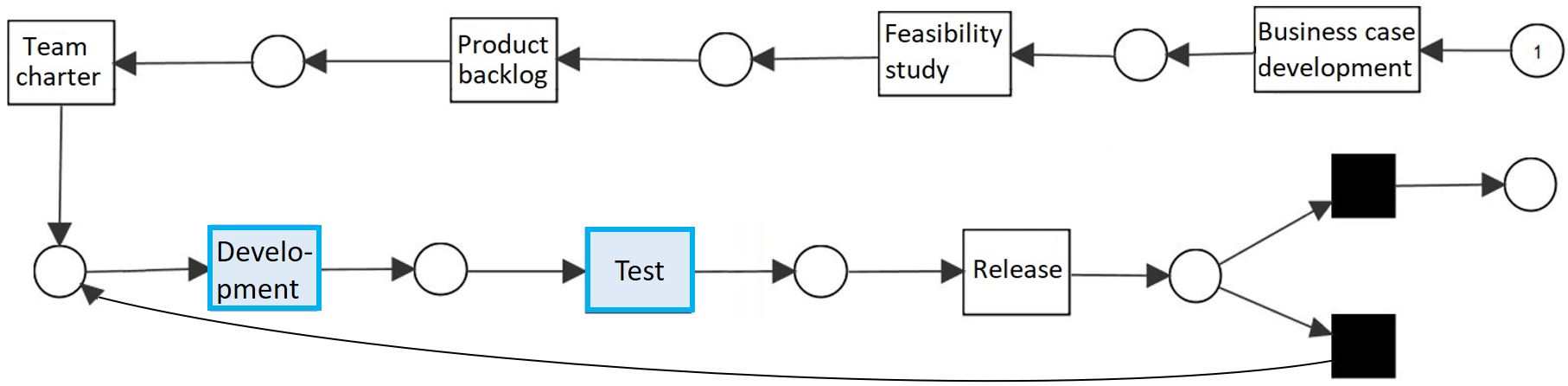}
	\caption{The Petri net model of the process of IT company described in Section \ref{sec::ex}.}\label{pic::ex}
\end{figure}

As the running example, we use an imaginary IT company that implements software for its customers. However, they do not do the maintenance of the released software. Here, each process instance is corresponding to the process of implementing one software. This process involves the following activities: business case development, feasibility study, product backlog, team charter, development, test, and release. The Petri-net model of this company is shown in Figure \ref{pic::ex}. We refer to the sub-model including two transitions ``development" and ``test" (the two blue activities in Figure \ref{pic::ex}) as \emph{implementation phase}.

The manager of the company is concerned about the duration of the implementation phase of projects. She wants to know what features determine the implementation phase duration. And also, if there is any way to reduce the implementation phase duration. If so, what would be the effect of changing each feature. These are valid questions to be asked before planning for re-engineering and enhancing the process. The manager believes that the following features of a project are the process features that might have a causal effect on its \emph{``implementation phase duration"} (the duration of implementation phase):
\begin{itemize}
    \item \emph{``Priority"} which is an attribute of business case development indicating how urgent the software is for the customer,
    \item \emph{``Team size"} which is an attribute of team charter indicating the number of resources working on a project,
    \item \emph{``Duration"} of product backlog activity, an attribute of product backlog, which indicates the duration of the product backlog activity.
\end{itemize} 
Analyzing the historical data from the company shows that there is a high correlation between every one of the three mentioned features and the duration of the implementation phase. We consider \emph{``Complexity"} (the complexity and hardness of the project) as another feature that is not recorded in the event log but has a causal effect on the duration of the implementation phase.

\begin{wrapfigure}{r}{7.5cm}
\includegraphics[width=7.5cm]{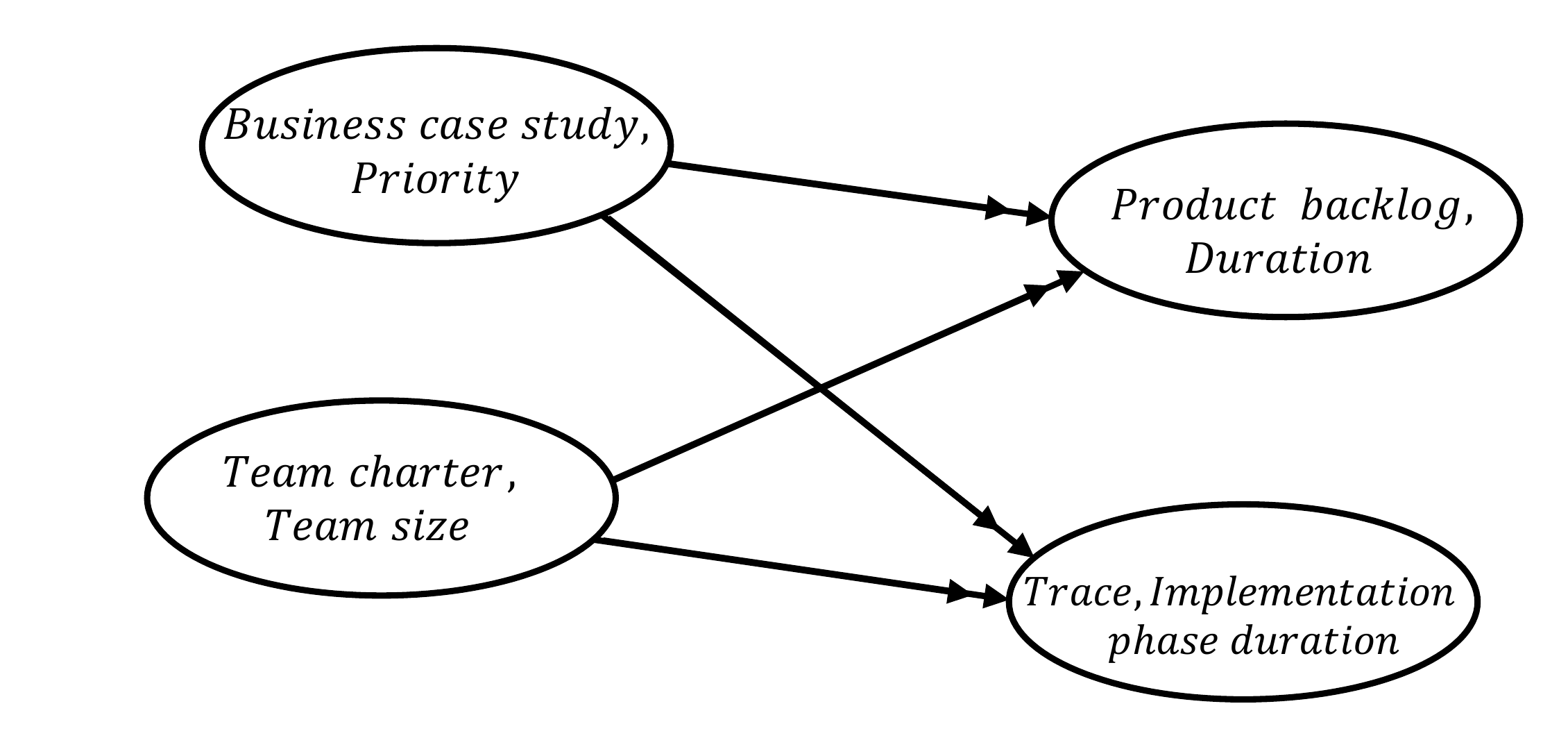}
\caption{A possible causal structure for the IT company.}\label{cg1}
\end{wrapfigure}

\begin{wrapfigure}{r}{7.5cm}
\includegraphics[width=7.5cm]{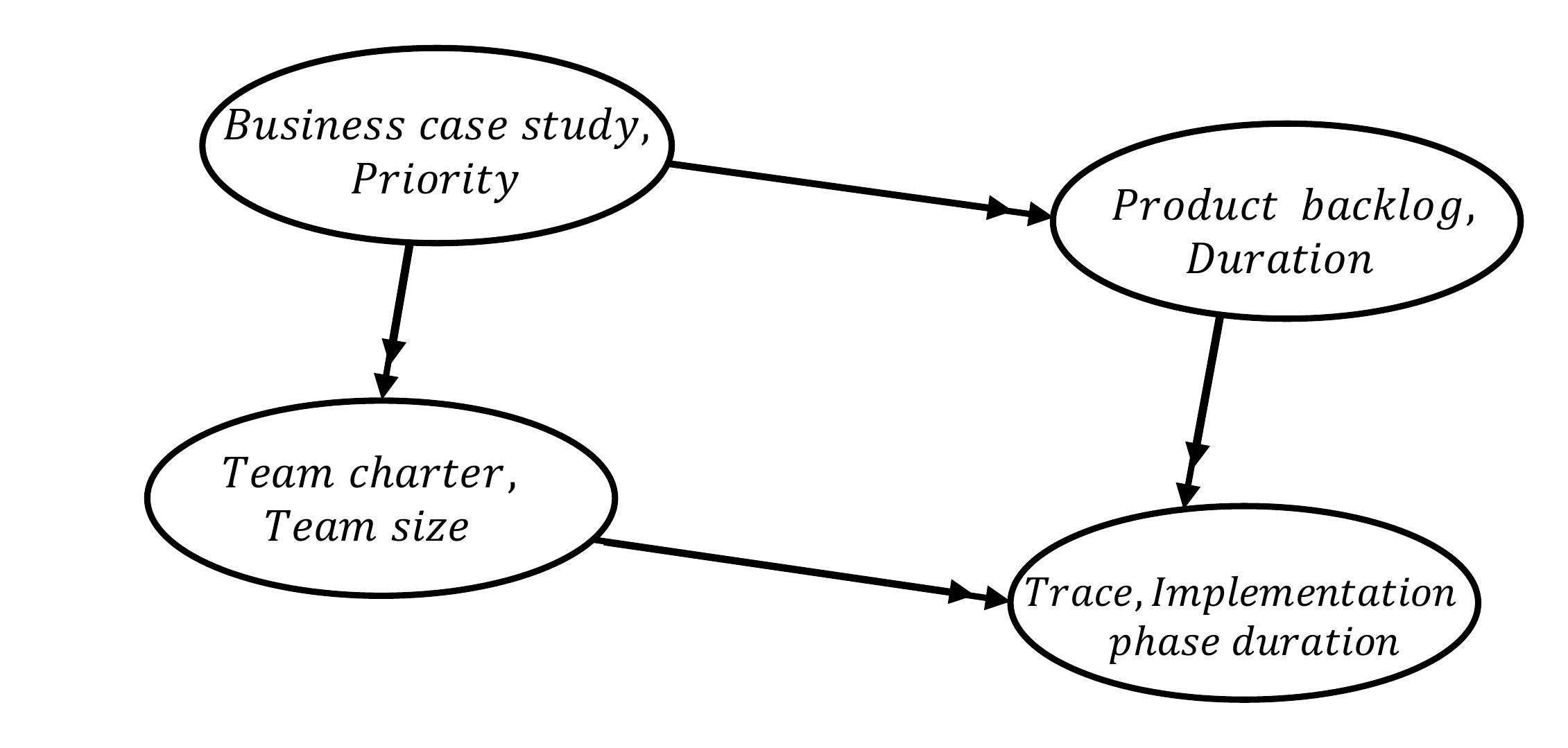}
\caption{A possible causal structure for the IT company.}\label{cg2}
\end{wrapfigure}

\begin{wrapfigure}{r}{7.5cm}
\includegraphics[width=7.5cm]{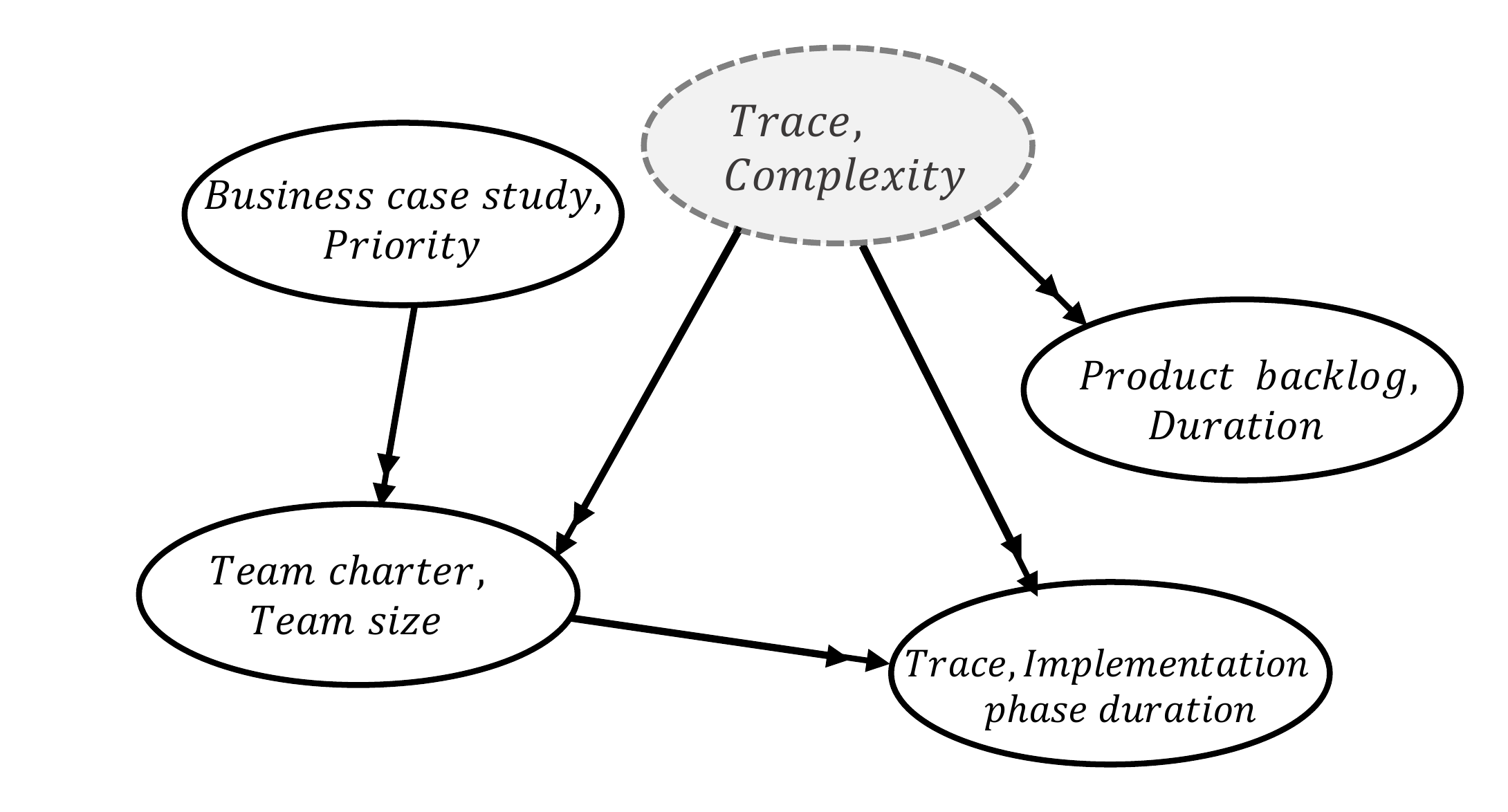}
\caption{A possible causal structure for the IT company.}\label{cg3}
\end{wrapfigure}

The structure of the causal relationship among the features has a high impact on the answers to the mentioned questions. In Figures \ref{cg1}, \ref{cg2}, and \ref{cg3}, three possible structures of the causal relationship among the features of the IT company are depicted\footnote{In these three figures and other figures in this paper that visualize networks of feature, the labels of the nodes are either of the form \emph{Trace, Attribute name} if the attribute name is related to a trace-level attribute, or of the form \emph{Activity name, Attribute name} if the attribute name is related to an event-level attribute. In the former case, the activity name indicated the activity that the attribute belongs to.}.

 According to Figure \ref{cg1}, just team size and priority have a causal effect on the duration of the implementation phase. But product backlog duration does not have any causal effect on the duration of the implementation phase even though they are highly correlated. Consequently, changing product backlog duration does not have any impact on the duration of the implementation phase.

According to Figure \ref{cg2}, all three features priority, product backlog duration, and team size influence the duration of the implementation phase. Thus, by changing each of these three features, one can influence the duration of the implementation phase.

Based on \ref{cg3}, we can conclude that the complexity, which is a hidden feature in the model (depicted by the gray dashed oval in Figure \ref{cg3}), causally influences both implementation phase duration and product backlog duration. Subsequently, the correlation among them is because of having a common cause. Grounded in this causal structure, it is not possible to influence the duration of the implementation phase by forcing product backlog activity to take place in a shorter or longer amount of time. 

It is worth noting that not all the features are actionable, i.e., in reality, it is not possible to intervene on some of the features. For example, in the mentioned IT company, we can imagine that the manager intervenes on team size by assigning more or fewer people to a project; but he cannot intervene in the complexity of a project. Judging whether a feature can be intervened requires using common sense and domain knowledge.

In the rest of this paper, we show how to answer such questions posed by the company manager. We first mention how to extract data in a meaningful way regarding the target feature (implementation phase duration in this example) and then we show how to discover the causal relationships between the process features and the structural equation model of the features that may affect the target feature using our method. Finally, we demonstrate how questions related to investigating the effect of intervention on the target feature can be answered in this framework. In Section \ref{synth}, we show the results of applying our method for answering the mentioned question by the IT company manager in this example.

\section{Related Work}\label{rw}
In the literature, there is plenty of work in the area of process mining dedicated to finding the root causes of a performance or compliance problem. The root cause analysis approach of the proposed methods usually involves classification \cite{GuptaAS15,de2016general}, and rule mining \cite{saniABG17}. The main problem of these approaches is that the findings of these methods are based on correlation which does not necessarily imply causation. 

The theory of causation based on the structural causal model has been studied deeply \cite{peters2017elements}. Also, a variety of domains benefit from applying methods from this domain(e.g. \cite{10.1145/3351095.3372850,abs-1808-06581}). However, there is little work on the application of the theory of causality in the area of process mining. There are some works in process mining that use causality theory. These includes: 
\begin{itemize}
    \item In \cite{hompes2017}, the authors propose an approach for discovering causal relationships between a range of business process characteristics and process performance indicators based on time-series analysis. The idea is to generate a set of time-series using the values of performance indicators, and then applying Granger causality test on them, to investigate and discover their causal relationships. Granger test is a statistical hypothesis test to detect predictive causality; consequently, the causal relationships using this approach might not be true cause-and-effect relationships. 
    \item In \cite{narendar2019}, the authors use the event log and the BPMN model of a process to discover the structural causal model discovery of the features of the process. They first apply loop unfolding on the BPMN model of the process and generate a partial order of features. They use the generated partial order to guide the search algorithm. In this work, it is assumed that the BPMN model of a process is its accurate model, which is not always the case.
\end{itemize}

There is also some work devoted to the case level root cause analysis \cite{bozorgi2020,qafari2021case}.

It is worth mentioning that all the above-mentioned approaches are based on statistical tests for discovering causal relationships. Consequently, these approaches are not feasible when there are a huge number of features. However, none of them provides a method for feature recommendation. Yet, there is some work on influence analysis that aims at finding such a set of features \cite{lehto2020discovering,lehto2016focusing,lehto2017focusing}.

\section{Preliminaries} \label{prel} \label{sec::preliminaries}
In this section, we describe some of the basic notations and concepts of the process mining and causal inference theory.

In the following section, we follow two goals: first, we describe the basic notations and concepts of the process mining and second, we show the steps involved in converting a given event log into a situation feature table.
\subsection{Process Mining}
Process mining techniques start from an \emph{event log} extracted from an information system. The atomic building block of an event log is an \emph{event}. An event indicates that an activity happened at a specific point in time for a specific case. A set of events that are related to a specific process instance are called a \emph{trace}. We can look at an event log as a collection of traces. An event log may include three different levels of attributes: log-level attributes, trace-level attributes, and event-level attributes. In the following, we explicitly define an event, trace and event log in a way that reflects reality and at the same time is suitable for our purpose. But first, we need to define the following universes and functions:
\begin{itemize}
    \item $\mathcal{U}_{att}$ is the universe of \emph{attribute names}, where $\{ actName, timestamp, caseID\} \subseteq \mathcal{U}_{att}$. $actName$ indicates the activity name, $timestamp$ indicates the timestamp of an event, and $caseID$ indicates that the event belongs to which case.
    \item $\mathcal{U}_{val}$ is the universe of \emph{values}. 
    \item $\val \in  \mathcal{U}_{att} \mapsto \mathbb{P}(\mathcal{U}_{val})$ as a function that returns the set of all possible values of a given attribute name\footnote{In this paper, it is assumed that the reader is familiar with sets, multi-sets, and functions. $\mathbb{P}(X)$ is the set of non-empty subsets of set $X \neq \emptyset$. Let $X$ and $Y$ be two sets. $f: X \not \mapsto Y$ is a partial function. The domain of $f$ is a subset of or equal to $X$ which is denoted by $dom(f)$. We write $f(x)= \bot$ if $x \not\in dom(f)$.}. 
    \item $\mathcal{U}_{map}=\{ m \in \mathcal{U}_{att} \not\mapsto \mathcal{U}_{val}|\forall at \in dom(m):m(at) \in \val(at) \}$ the universe of all mappings from a set of attribute names to attribute values of the correct type.
\end{itemize}

Also, we define $\bot$ as a member of $\mathcal{U}_{val}$ such that $\bot \not\in \val (at)$ for all $ at \in \mathcal{U}_{att}$. We use this symbol to indicate that the value of an attribute is unknown, undefined, or is missing.

Now, we define an event as follows:
\begin{definition}[Event]\label{def:event}
	An \emph{event} is an element of $e \in \mathcal{U}_{map} $, where $e(actName) \neq \bot$, $e(timestamp)\neq \bot$, and $e(caseID)\neq \bot$. We denote the universe of all possible events by  $\mathcal{E}$ and the set of all non-empty chronologically ordered sequences of events that belong to the same case (have the same value for $caseID$) by $\mathcal{E}^+$. If $\langle e_1, \dots , e_n\rangle \in \mathcal{E}^+$, then for all $1\leq i<j\leq n$, $e_i(timestamp) \leq e_j(timestamp) \wedge e_i(caseID) = e_j(caseID)$. 
\end{definition} 

\begin{example}\label{ex:event}
	The events in the following table are some of the possible events for the IT company in Section \ref{sec::ex}.
	\begin{center}
		\resizebox{\textwidth}{!}{
		\begin{tabular}{ l }
			
			\hline

$e_1:=\{(caseID,1), (actName,\text{``Business case development"}), (timestamp, t_1), (Priority, 2)\}$\\ $e_2:=\{(caseID,1),(actName,\text{``Feasibility study"}),(timestamp, t_2)\}$\\ 
            $e_3:=\{(caseID,1), (actName,\text{``Product backlog"}),(timestamp, t_3),  ( Duration,35)\}$  \\
			$e_4:=\{(caseID,1), (actName,\text{``Team charter"}),(timestamp, t_4), (team\ size,21)\}$ \\
			$e_5:=\{(caseID,1),(actName,\text{``Development"}),(timestamp, t_5), ( Duration,200) \}$ \\ $e_6:=\{(caseID,1), (actName,\text{``Test"}),(timestamp, t_6), (Duration,79) \}$  \\
			$e_7:=\{(caseID,1), (actName,\text{``Release"}),(timestamp, t_7) \}$ \\   
			$e_8:=\{(caseID,2), (actName, \text{``Business case development"}) ,(timestamp, t_8), (Priority, 1) \}$  \\
			$e_9:=\{(caseID,2), (actName, \text{``Feasibility study"}), (timestamp, t_9)\}$ \\
		$e_{10}:=\{(caseID,2), (actName, \text{``Product backlog"}),(timestamp, t_{10}), ( Duration,63) \}$ \\ 
		$e_{11}:=\{(caseID,2), (actName,\text{``Team charter"}),(timestamp, t_{11}), (team\ size,33)\}$ \\
		$e_{12}:=\{(caseID,2), (actName,\text{``Development"}), (timestamp, t_{12}), (Duration,226) \}$  \\
		$e_{13}:=\{(caseID,2), (actName, \text{``Test"}),(timestamp, t_{13}), ( Duration,74) \}$ \\   $e_{14}:=\{(caseID,2), (actName,\text{``Release"}),(timestamp, t_{14})\}$\\
		$e_{15}:=\{(caseID,2), (actName, \text{``Development"}), (timestamp, t_{15}), (Duration,62)\}$  \\
		 $e_{16}:=\{(caseID,2), (actName,\text{``Test"}), (timestamp, t_{16}),(Duration,117) \}$ \\ 
		 $e_{17}:=\{(caseID,2), (actName,\text{``Release"}), (timestamp, t_{17})\}$  \\
			\hline
		\end{tabular}}
	\end{center}

\end{example}
Each event may have several attributes which can be used to group the events. For $at \in \mathcal{U}_{att}$, and $V \subseteq \val (at)$, we define a group of events as the set of those events in $\mathcal{E}$ that assign a value of $V$ to the attribute $at$; i.e. $$group(at, V) = \{e \in \mathcal{E}| e(at) \in V\}.$$

Some of the possible groups of events are: 
\begin{itemize}
    \item the set of events with specific activity names,
    \item the set of events which are done by specific resources,
    \item the set of events that start in a specific time interval during the day, or,
    \item the set of events with a specific duration.
\end{itemize}
 We denote the universe of all event groups by $\mathcal{G} = \mathbb{P} (\mathcal{E})$.
\begin{example} \label{ex:group}
Here are some possible event groups based on the IT company in Section \ref{sec::ex}.
\begin{align*}		       
    &G_1 \coloneqq group(actName, \{ \text{``Business case development"}\}) \\
    &G_2 \coloneqq group(actName, \{ \text{``Product backlog"}\}) \\
	&G_3 \coloneqq group(actName, \{ \text{``Team charter"}\})\\
	&G_4 \coloneqq group(actName, \{ \text{``Development"}\})\\
	&G_5 \coloneqq group(team\ size, \{ 33, 34, 35\})
\end{align*}	
\end{example}

Based on the definition of an event, we define an event log as follows:
\begin{definition}[Event Log]\label{def:eventLog}
We define the universe of all event logs as $\Ul = \mathcal{E}^+ \not\mapsto \mathcal{U}_{map}$. Let $L$ where $L \in \Ul$ be an event log, we call each element $(\sigma , m) \in L$ a \emph{trace}.
\end{definition}
One of our assumptions in this paper is the  uniqueness of events in event logs; i.e., given an event log $L \in \Ul$, we have $\forall (\sigma_1,m_1), (\sigma_2,m_2) \in L: e_1 \in \sigma_1 \wedge e_2 \in \sigma_2 \wedge e_1 = e_2 \implies (\sigma_1,m_1)= (\sigma_2,m_2)$ and $\forall (\langle e_1, \dots , e_n\rangle , m) \in L : \forall 1\leq i <j \leq n : e_i \neq e_j$. This property can easily be ensured by adding an extra identity attribute to the events. 

Also, we assume that the uniqueness of the ``caseID" value for traces in a given event log $L$. In other words, $\forall (\sigma_1,m_1), (\sigma_2,m_2) \in L: e_1 \in \sigma_1 \wedge e_2 \in \sigma_2 \wedge e_1(caseID) = e_2(caseID) \implies (\sigma_1,m_1)= (\sigma_2,m_2)$.
\begin{example}\label{ex:traceLog}
	$L_{IT}= \{ \lambda_1, \lambda_2\}$ is a possible event log for the IT company in \ref{sec::ex}. $L_{IT}$ includes two traces $\lambda_1$ and $\lambda_2$, where:
	\begin{itemize}
	    \item $\lambda_1 \coloneqq (\langle e_1,\dots e_7\rangle, \{ (Responsible, Alice) \})$ and
	    \item $\lambda_2 \coloneqq (\langle e_8,\dots e_{17}\rangle, \{ (Responsible, Alex)\} )$.
	\end{itemize}
	Here $t_1,\dots , t_{17}$ are unique timestamps where $t_1<\dots <t_7$ and $t_8<\dots<t_{17}$.
\end{example}

As a preprocessing step, we enrich the event log by adding many derived features to its traces and events. There are many different derived features related to any of the process perspectives; the time perspective, the data flow-perspective, the control-flow perspective, the conformance perspective, or the resource/organization perspective of the process. We can compute the value of the derived features from the event log or possibly other sources. 

Moreover, we can enrich the event log by adding aggregated attributes to its events and traces. Let $L \in \Ul$ be an event log, $k \in \mathbb{N}$ (a non-zero natural number) the number of time windows, $t_{min}$ the minimal timestamp, and $t_{max}$ the maximum timestamp in $L$, we divide the time span of $L$ into $k$ consecutive time windows with equal length (the length of each time window is $(t_{max} - t_{min})/k$ and compute the value of aggregated attributes for each of these $k$ time windows. In other words, We define $\xi: \Ul \times \mathcal{U}_{att} \times \mathbb{N} \times \val (timestamp) \to \mathbb{R}$ as a function that given an event log, an aggregated attribute name, the number of time windows, and a timestamp returns the value of the given aggregated attribute in the time window that includes the timestamp. We can use $\xi$ for aggregated attributes at both the event and the  trace-level. More precisely, given $L \in \Ul$, $(\sigma, m) \in L$, $e \in \sigma$, $k \in \mathbb{N}$, and $at \in \mathcal{U}_{att}$ where $at$ is an aggregated attribute, we define $e(at) = \xi(L, at, k, e(timestamp))$ and $m(at) = \xi(L, at, k, t')$ where $t' = max \{ e(timestamp) |e \in \sigma\} $. Some of the possible aggregated attributes are: the number of waiting customers, workload (in the process-level), average service time, average waiting time (in the trace and event-level), number of active events with a specific activity name, number of waiting events with a specific activity name (in the event-level), average service time, average waiting time (in the resource-level).



While extracting the data from an event log, we assume that the event recording delays by the information system of the process were negligible. Considering the time order of cause and effect, we have that only the features that have been recorded before the occurrence of a specific feature can have a causal effect on it. So the relevant part of a trace to a given feature is a prefix that trace, which we call such a prefix of a trace a \emph{situation}. Let $\prfx(\langle e_1,\dots , e_n \rangle ) = \{\langle e_1,\dots , e_i \rangle | 1\leq i \leq n \} $, a function that returns the set of non-empty prefixes of a given sequence of events. Using $\prfx$ function we define a situation as follows:
\begin{definition}[Situation]\label{def::situation}
	 We define $\mathcal{U}_{situation} = \mathcal{E}^+ \times \mathcal{U}_{map}$ as the universe of all situations. We call each element $(\sigma, m) \in \mathcal{U}_{situation}$ a \emph{situation}. Considering $L \in \Ul$, we define the set of situations of $L$ as $S_L = \{(\sigma, m) | \sigma \in prfx (\sigma ') \wedge (\sigma', m) \in L \}$.
\end{definition}

Among the possible subsets of $S_L$ of a given event log $L$, we distinguish two important type situation subsets of $S_L$. The first type is the \emph{$G$-based situation subset} of $L$ where $G \in \mathcal{G}$ and includes those situations in $S_L$ that their last event (the event with maximum timestamp) belongs to $G$. The second type is the \emph{trace-based situation subset}, which includes the set of all traces of $L$.
\begin{definition}[Situation Subset]
     Given $L \in \mathcal{L}$ where $S_L \subseteq \mathcal{U}_{situation}$ is the set of situations of $L$, and $G \in \mathcal{G}$, we define
     \begin{itemize}
         \item \emph{$G$-based situation subset} of $L$ as $S_{L,G}= \{ (\langle e_1, \dots , e_n \rangle, m) \in S_L| e_n \in G\}$, and
         \item \emph{trace-based situation subset} of $L$ as $S_{L,\bot} =L$.
     \end{itemize} 
\end{definition}

\begin{example}\label{ex:situation}
Three situations $s_1$, $s_2$, and $s_3$, where $s_1, s_2, s_3 \in S_{L_{IT},G_4}$ ($G_4$ in Example \ref{ex:group}, generated using the trace mentioned in Example \ref{ex:traceLog} are as follows:
\begin{align*}
    &s_1 \coloneqq (\langle e_1,\dots e_5\rangle, \{ (Responsible, Alice) \})\\
    &s_2 \coloneqq (\langle e_8,\dots e_{12}\rangle, \{ (Responsible, Alex) \})\\
    &s_3 \coloneqq (\langle e_8,\dots e_{15}\rangle, \{ (Responsible, Alex)\} )
\end{align*}
Note that $G_4 \coloneqq group(actName, \{ \text{``Development"}\})$ and we have $\{e_5,e_{12},e_{15}\} \subseteq G_4$. In other words $e_5(actName) = e_{12}(actName) = e_{15}(actName) = \text{``Development"}$.
\end{example}

If a process includes decision points, then one of the derived attributes that can be added to the event log when enriching the event log is the \emph{choice} attribute. A choice attribute is added to the activity that happens before the decision point and its value indicates which activity has been enabled as the result of the decision that has been made. So we can use an added choice attribute and its values to group the events in an event log and extract a situation subset based on the occurrence of that specific choice. We already defined two important types of situation subsets; group-based situation subsets and trace-based situation subsets. We also distinguish the \emph{choice-based situation subsets} where the situation subset is extracted based on events that have a specific choice attribute. These situation subsets are important as they are conceptually related to a decision point.

When extracting the data, we need to distinguish trace-level attributes from event-level attributes. We do that by using \emph{situation features} which is identified by a group of events, $G$ (possibly $G =\bot$), and an attribute name, $at$. Each situation feature is associated with a function defined over the situations. This function returns the proper value for the situation feature regarding $at$ and $G$ extracted from the given situation. More formally:
\begin{definition}[Situation Feature]\label{def:sf}
	We define $\Usf =\mathcal{U}_{att}\times (\mathcal{G} \cup \{\bot\})$ as the universe of the \emph{situation features}. Each situation feature is associated with a function $\sfunc_\sitf : \mathcal{U}_{situation}\not \mapsto \mathcal{U}_{val}$ such that given $\sitf =(at, G)$ where $at \in \mathcal{U}_{att}$, and $G \in \mathcal{G} \cup \{\bot \}$ 
	\begin{itemize} we have:
	    \item if $G=\bot$, then $\sfunc_{(at,G)} ((\sigma,m)) = m(at) $ and
	    \item if $G \in \mathcal{G}$, then $\sfunc_{(at,G)} ((\sigma,m)) = e(at) $ where $e =\displaystyle\argmax_{\substack{e' \in G \cap \{ e" \in \sigma \}}}e'(timestamp) $ for $(\sigma , m)\in \mathcal{U}_{situation}$.
	\end{itemize}
	 We denote the universe of the situation features as $\Usf$.
\end{definition}
We can consider a situation feature as an analogy to the feature (a variable) in a tabular data. Also, we can look at the corresponding function of a situation feature as the function that determines the mechanism of extracting the value of the situation feature from a given situation.
Given a situation $(\sigma, m)$ and a situation feature $(at,G)$, if $G =\bot$, its corresponding function returns the value of $at$ in trace level (i.e., $m(at)$). However, if $G \neq \bot$, then the function returns the value of $at$ in $e\in \sigma$ that belongs to $G$ and happens last (has the maximum timestamp) among those events of $\sigma$ that belong to $G$. 
\begin{example}\label{ex:sf}
We can define the following situation features using the information mentioned in the previous examples:
\begin{equation*}
\begin{aligned}[c]
&\sitf_1 \coloneqq  (Team\ size, G_3) \\
&\sitf_2 \coloneqq (Duration, G_2)\\
&\sitf_5 \coloneqq  (Implementation\ phase\ duration, \bot).
\end{aligned}
\ \ \ 
\begin{aligned}[c]
&\sitf_3 \coloneqq  (Priority, G_1)\\
&\sitf_4 \coloneqq (Duration, G_4) \\
\end{aligned}
\end{equation*}
Also, considering $s_1$ (Example \ref{ex:situation}), we have:
\begin{equation*}
\begin{aligned}[c]
&\sfunc_{\sitf_1}(s_1) =21\\
&\sfunc_{\sitf_2}(s_1)= 35\\
&\sfunc_{\sitf_5}(s_1)= 279
\end{aligned}
\ \ \ 
\begin{aligned}[c]
&\sfunc_{\sitf_3}(s_1)= 2\\
&\sfunc_{\sitf_4}(s_1)= 200\\
\end{aligned}
\end{equation*}
where $s_1$ is one of the situations mentioned in \ref{ex:situation}.
\end{example}

We interpret a nonempty set of situation features, which we call it a \emph{situation feature extraction plan}, as an analog to the schema of tabular data. More formally;
\begin{definition}[Situation Feature Extraction Plan]\label{def:sfep}
   We define a \emph{situation feature extraction plan} as $\SF \subseteq \Usf$ where $\SF \neq~\emptyset$.
\end{definition}
 
\begin{example}
A possible situation feature extraction plan for the IT company in Section \ref{sec::ex} is as follows:
\begin{align*}
  \SF_{IT} =& \{(Team\ size, G_3), (Duration, G_2), (Priority, G_1), (Duration, G_4)\} \\=  &\{\sitf_1, \sitf_2, \sitf_3, \sitf_4\}.  
\end{align*}
\end{example}

We can map each situation to a data point according to a given situation feature extraction plan. We do that as follows:

\begin{definition}[Instance]	\label{def::instance}
Let $s \in \mathcal{U}_{situation}$ and $\SF \subseteq \Usf$ where $\SF \neq \emptyset$. We define the \emph{instance} $inst_{\SF} (s)$ as $inst_{\SF} (s) \in \SF \to \mathcal{U}_{val}$ such that $\forall \sitf \in \SF : (inst_{\SF} (s) )(\sitf) = \sfunc_\sitf (s)$. We denote the universe of all possible instances as: 
	\begin{equation*}
	    \begin{split}
	        	\Uinst =\bigcup_{s \in  \mathcal{U}_{situation}} \bigcup_{\substack{
\SF \subseteq \Usf \\
\SF \neq \emptyset
}} \{ inst_{\SF} (s) \} .
	    \end{split}
	\end{equation*}
\end{definition}

An instance is a set of pairs where each pair is composed of a situation feature and a value. With a slight abuse of notation, we define $\val(\sitf) = \val(at)$ where $\sitf = (at, G)$ is a situation feature.
\begin{example}\label{ex:instance}
Considering $\SF_{IT}$ from Example \ref{ex:sf} and the situations from Example \ref{ex:situation}. We have:
\begin{align*}
    inst_{\SF_{IT}} (s_1)=&\{ ((Team\ size, G_3),21), ( (Duration, G_2),35), ( (Priority, G_1), 2),\\ & ((Duration, G_4), 200) \} = \{(\sitf_1 , 21) , (\sitf_2 , 35), (\sitf_3, 2), (\sitf_4, 200)  \} \\
    inst_{\SF_{IT}} (s_2)=&\{ ((Team\ size, G_3),33), ( (Duration, G_2),63), ( (Priority, G_1), 1),\\&((Duration, G_4), 226) \} = \{(\sitf_1 , 33) , (\sitf_2 , 63), (\sitf_3, 1), (\sitf_4, 226)  \} \\
    inst_{\SF_{IT}} (s_3)=&\{ ((Team\ size, G_3),33), ( (Duration, G_2),63), ( (Priority, G_1), 1),\\&((Duration, G_4), 62) \} = \{(\sitf_1 , 33) , (\sitf_2 , 63), (\sitf_3, 1), (\sitf_4, 62)  \} \\
\end{align*}
\end{example}

Given a situation feature extraction plan $\SF$, we consider one of its situation features as the class situation feature, denoted as $\csf$ and $\SF \setminus \{\csf \}$ as descriptive situation features. Given $\SF \subseteq \Usf$, $\csf \in \SF$ where $\csf = (at,G)$, and an event log $L$, we can generate a class situation feature sensitive tabular data-set. We call such a tabular data set a \emph{situation feature table}. To do that, we first generate $S_{L,G}$ and then we generate the situation feature table which is the bag of instances derived from the situations in $S_{L,G}$, regarding $\SF$. Note that choosing $S_{L,G}$ such that $G$ is the same group as the one in class situation feature (where we have $\csf = (at,G)$), ensures the sensitivity of the extracted data to the class situation feature.  More formally;

\begin{definition}[Situation Feature Table] \label{def::table}
Let $L \in \Ul$ be an event log, $\SF\subseteq \Usf$ a situation feature extraction plan, and $\csf = (at,G) \in \SF$. We define a \emph{situation feature table} $T_{L,\SF,(at,G)}$ (or equivalently $T_{L,\SF,\csf}$) as follows:
$$T_{L,\SF,(at,G)} = [inst_{\SF}(s)| s \in S_{L,G}].$$
\end{definition}
Note that if $\csf = (at, G)$ where $G \in \mathcal{G}$, then the situation feature table $T_{L,\SF,\csf}$ includes the instances derived from the situations in $G$-based situation subset $S_{L,G}$. However, if $G = \bot$, then it includes the situations derived from the situations in trace-based situation subset $S_{L,\bot}$.

\begin{example} \label{ex:table}
Based on Example \ref{ex:instance} we have $$T_{L_{IT},\SF_{IT},(\text{Duration}, G_4)}= [ inst_{\SF_{IT}} (s_1), inst_{\SF_{IT}} (s_2), inst_{\SF_{IT}} (s_3)].$$ Note that in this example, the class situation feature is $\csf = \sitf_4 = (\text{Duration}, G_4)$. Another way to present $T_{L_{IT},\SF_{IT},(\text{Duration}, G_4)}$ is as follows:
\begin{center}
	\resizebox{10cm}{!}{
		\begin{tabular}{ |c | c| c| c| }
			\hline
		    $\sitf_1 = (Team\ size, G_3)$&$\sitf_2 = (Duration, G_2)$&$\sitf_3 = (Priority, G_1)$&$\sitf_4 = (Duration, G_4)$\\
		    \hline
		    21 & 35 & 2 & 200\\
            33 & 63 & 1 & 226 \\
            33 & 63 & 1 & 117\\
            \hline
	\end{tabular}}\label{tab:ex:table}
\end{center} 
In this table, the first row is corresponding to the $inst_{\SF_{IT}} (s_1$), the second row is corresponding to the $inst_{\SF_{IT}} (s_2)$, and the third row is corresponding to the $inst_{\SF_{IT}} (s_3)$.
\end{example}

\subsection{Structural Equation Model}
A structural equation model is a data generating model in the form of a set of equations. Each equation encodes how the value of one of the situation features is determined by the value of other situation features. It is worth noting that these equations are a way to determine how the observational and the interventional distributions are generated and should not be considered as normal equations. More formally\footnote{Definition \ref{SEM} and \ref{intervention} are based on \cite{peters2017elements}.};

\begin{definition}[Structural Equation Model (SEM)]\label{SEM}
    Let $T_{L,\SF,\csf}$ be a situation feature table, in which $L \in \Ul$, $\SF \subseteq \Usf$, and $\csf \in  \SF$. The SEM of $T_{L,\SF,\csf}$ is defined as $\eq \in \SF \to Expr(\SF)$ where for each $\sitf \in SF$, $Expr (\sitf)$ is an expression of the situation features in $\SF \setminus \{ \sitf \}$ and possibly some noise $N_{\sitf}$. It is needed that the noise distributions $N_{\sitf}$ of $\sitf \in \SF$ be mutually independent.
\end{definition}
 We need $\SF$ to be \emph{causal sufficient}, which means $\SF$ includes all relevant situation features. Based on Definition \ref{SEM}, given a SEM $\eq$ over the $\SF$ of a situation feature table $T_{L,\SF,\csf}$, for each $\sitf \in \SF$, the right side of expression $\sitf = Expr(\SF)$ in $\eq$ does not include $\sitf$. 

Given $\eq$ over the $\SF$ of a situation feature table $T_{L,\SF,\csf}$, the \emph{parents} of the $\sitf \in \SF$ is the set of situation features that appear in the right side of expression $\eq (\sitf)$. The set of parents of a situation feature includes those situation features with a direct causal effect on it. 
\begin{example} \label{ex:sem}
A possible SEM for the situation feature table mentioned in Example \ref{ex:table} is as follows:
\begin{center}
	\resizebox{11cm}{!}{
		\begin{tabular}{ l l }
			$(Priority, G_1) = N_{(Priority, G_1)}$ &$N_{(Priority, G_1)}\sim Uniform (1,3)$  \\
			$(Team\ size, G_3) =10(Priority, G_1) +N_{(Team\ size, G_3)}$ &   $N_{(team\ size, G_3)} \sim Uniform(1,15)$ \\
			
			$(Duration, G_2) =2 (team\ size, G_3) + N_{(Duration, G_2)}$ & $N_{(Duration, G_2)} \sim Uniform(-5,5)$  \\

			$(Duration, G_4) =2(Duration, G_2) \times (Priority, G_1) $  &$N_{(Duration, G_4)} \sim Uniform(-100,100)$  \\
			$\ \ \ \ \ \ \ \ \ \ \ \ \ \ \ \ \ \ \ \ \ \ \ \ \ \ \ \ \  +  (team\ size, G_3) +N_{(Duration, G_4)}$& \\
	\end{tabular}}\label{tab:ex:sem}
\end{center} 
\end{example}

The structure of the causal relationships between the situation features in a SEM can be encoded as a directed acyclic graph, which is called \emph{causal structure}. Given a SEM $\eq$ on a set of situation features $\SF$, each vertex in its corresponding causal structure is analogous to one of the situation features in $\SF$. Let $\sitf_1,\sitf_2 \in \SF$, there is a directed edge from $\sitf_1$ to $\sitf_2$ if $\sitf_1$ appears in the right side of expression $\eq(\sitf_2)$. More formally,
\begin{definition}[Causal Structure]
    Let $\eq$ be the SEM of the situation feature table $T_{L,\SF,\csf}$. We define the corresponding \emph{causal structure} of $\eq$ as a directed acyclic graph $(\U, \twoheadrightarrow)$ where $\U = \SF$ and $(\sitf_1, \sitf_2) \in \twoheadrightarrow $ if $\sitf_1, \sitf_2 \in \SF$ and $\sitf_1$ appears in the right side of expression $\eq(\sitf_2)$. 
\end{definition}
In the rest of this paper, we use $\sitf_1 \twoheadrightarrow \sitf_2$ instead of $(\sitf_1, \sitf_2) \in \twoheadrightarrow $.

\begin{figure}
 	\includegraphics[width=80mm]{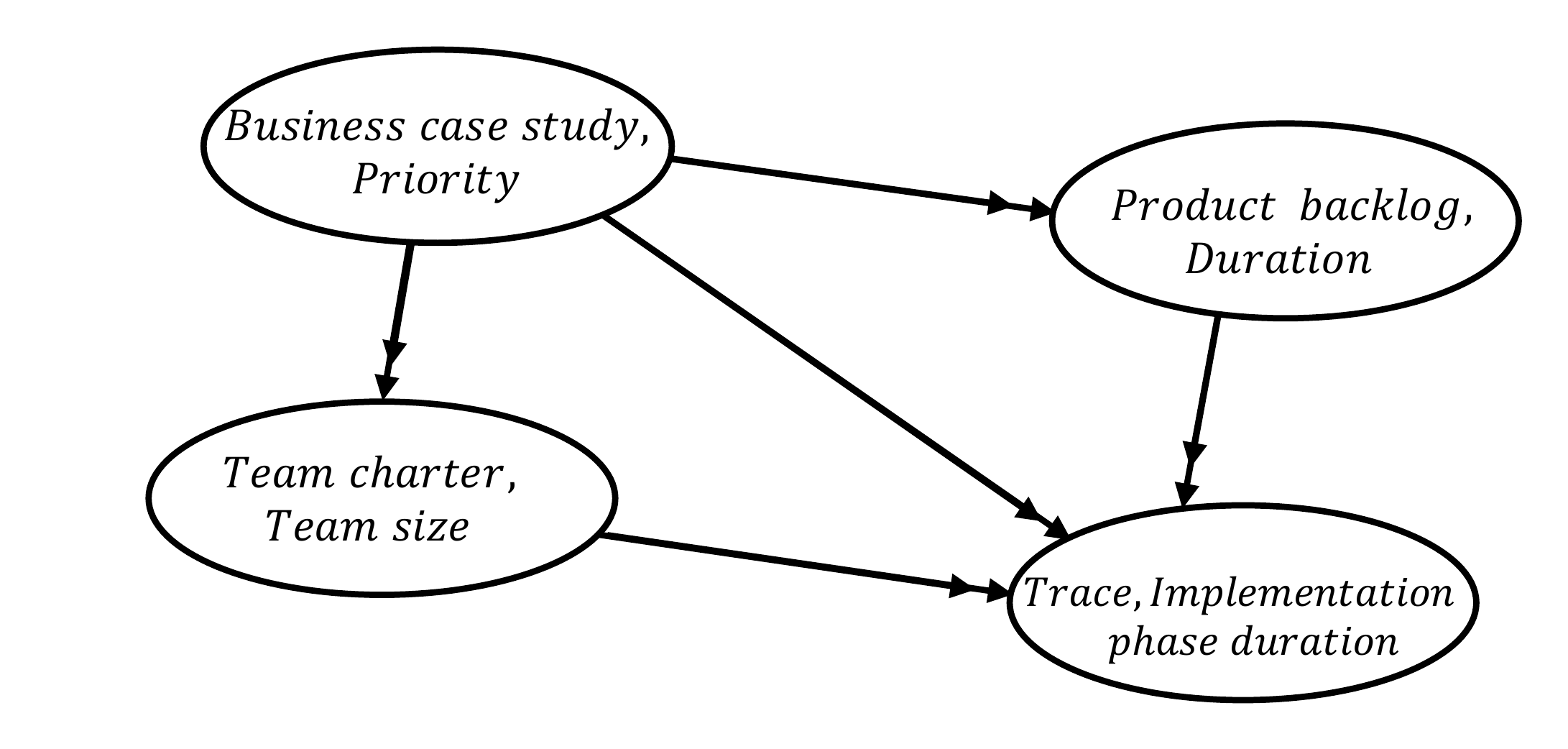}
 	\caption{The causal structure of the SEM mentioned in Example \ref{ex:sem}.}\label{fig:ex:causal Structure}
\end{figure}

Having a situation feature table $T_{L,\SF,\csf}$, the structural equation model of its situation features can be provided by a customer who possesses the process domain knowledge or in a data-driven manner. 
\begin{example}\label{ex:causalStructure}
The causal structure of the SEM mentioned in Example \ref{ex:sem} is as depicted in Figure \ref{fig:ex:causal Structure}.
\end{example}

To predict the effect of manipulating one of the situation features on the other situation features, we need to intervene on the SEM by actively setting the value of one (or more) of its situation features to a specific value (or a distribution). Here, we focus on atomic interventions where the intervention is done on just one of the situation features by actively forcing its value to be a specific value.

\begin{definition}[Atomic Intervention]\label{intervention}
	Given an SEM $\eq$ over $\SF$ where $\sitf \in \SF \setminus \{\csf \}$, and $c \in \val (\sitf)$, the SEM $\eq'$ after the intervention on $\sitf$ is obtained by replacing $\eq(\sitf)$ by $ \sitf = c$ in $\eq$. 
\end{definition}

Note that the corresponding causal structure of $\eq'$ (after intervention on $\sitf$) is obtained from the causal structure of $\eq$ by removing all the incoming edges to $\sitf$ \cite{peters2017elements}. When we intervene on a situation feature, we just replace the equation of that situation feature in the SEM and the others do not change as causal relationships are autonomous under interventions \cite{peters2017elements}.
\begin{example}
We can intervene on the SEM introduced in Example \ref{ex:sem} by forcing the team size to be 13. For this case, the SEM under the intervention is as follows:
\begin{center}
	\resizebox{11cm}{!}{
		\begin{tabular}{ l l }
			$(Priority, G_1) = N_{(Priority, G_1)}$ &$N_{(Priority, G_1)}\sim Uniform (1,3)$  \\
			$(Team\ size, G_3) =13$ &   \\
			
			$(Duration, G_2) =2 (team\ size, G_3) + N_{(Duration, G_2)}$ & $N_{(Duration, G_2)} \sim Uniform(-5,5)$  \\

			$(Duration, G_4) =2(Duration, G_2) \times (Priority, G_1) $  &$N_{(Duration, G_4)} \sim Uniform(-100,100)$  \\
			$\ \ \ \ \ \ \ \ \ \ \ \ \ \ \ \ \ \ \ \ \ \ \ \ \ \ \ \ \  + (team\ size, G_3) +N_{(Duration, G_4)}$& \\
	\end{tabular}}\label{sem:sec:ex}
\end{center} 
\end{example}

\section{Approach} \label{app}
Observing a problem in the process, we need to find a set of situation features $\SF$ which not only include $\csf$ (the situation feature capturing the problem) but also be causal sufficient. The expressiveness of the discovered SEM is highly influenced by $\SF$ (even though SEMs, in general, can deal with latent variables). Considering the variety of the possible situation features captured by the event log and the derived ones, finding the proper set $\SF$ and also those values of the situation features (or combination of values) that contribute more to the problem is a complicated task and needs plenty of domain knowledge. 

We know that correlation does not mean causation. On the other hand, if a situation feature is caused by another situation feature (set of situation features), this implies that there is a correlation between the given situation feature and its parents. We use this simple fact for the automated situation feature recommendation. It is worth noting that the proposed automated situation feature recommendation method is one of the many possible choices. The automated situation feature recommendation method and the SEM discovery process are described in the following:

\subsection{Automated situation feature Recommendation}
Given an event log $L \in \Ul$ and the class situation feature $\csf = (at , G)$, we declare a nominal situation feature $\sitf$ as a possible cause of $\csf$ if there exists a value $v \in \val (\sitf )$ that appears in big enough portion (at least $\alpha$ where $0 < \alpha \leq 1$) of the situations of $S_{L,G}$ with the undesirable (problematic) result for $\csf$. When $\sitf$ is a numerical situation feature, we use equal width binning for the values of $\sitf$ that appear in $L$ where the number of bins, $b$, is given by the user. We consider $\sitf$ as a possible cause of $\csf$ if there exists a bin of values of $\sitf$ in $L$ such that the values of that bin appears in more than of $\alpha$ portion of situations in $S_{L,G}$ with the undesirable value for $\csf$. More formally:

\begin{definition}[Potential Causal situation feature]
Let $L \in \Ul$  be an event log, $\csf = (at , G)$ the class situation feature where $G \in \mathcal{G} \cup \{ \bot \}$, $\alpha$ a threshold where $0 < \alpha \leq 1$, and $\val(\csf)_{\downarrow}$ denotes the set of undesirable values of $\csf$. We consider a situation feature $\sitf$ a possible cause of $\csf$ if one of the following two conditions holds:

 If $\sitf$ is a nominal situation feature: 
    $$ \exists_{v \in \val (\sitf )}\frac{|\{s\in S_{L,G} | \sfunc_\sitf (s)= v \wedge \sfunc_\csf  (s) \in \val (\csf)_{\downarrow}\}| }{|\{s\in S_{L,G} | \sfunc_\csf (s) \in \val (\csf)_{\downarrow}\}|} \geq \alpha.$$
    
If $\sitf$ is a numerical situation feature: 
    $$ \exists_{0 \leq i \leq b-1 }\frac{|\{s\in S_{L,G} | \sfunc_\sitf (s)\in [\frac{i(v_{max}-v_{min}+1)}{b},\frac{(i+1)(v_{max}-v_{min}+1)}{b}) \wedge \sfunc_\csf  (s) \in \val (\csf)_{\downarrow}\}| }{|\{s\in S_{L,G} | \sfunc_\csf (s) \in \val (\csf)_{\downarrow}\}|} \geq \alpha$$
    where $b \in \mathbb{N}$ denotes the number of bins, $v_{max}$ is the maximum value and $v_{min}$ is the minimum value for $\sitf$ in $L$.

  Moreover, we define the set of all potential causes of $\csf$ as the set of all $\sitf \in \mathcal{U}_{\sitf}$ for which one of the above inequalities holds.
\end{definition}

We present the set of the potential causes to the user as a set of tuples $(\sitf,v)$ where $\sitf \in \Usf$ and $v \in \val (\sitf)$ (if $\sitf$ is a numerical situation feature, then $v$ is the lower bound of the bin) in the descending order regarding the portion of the situations of $S_{L,G}$ with the undesirable result that has value $v$ for $\sitf$ (has a value in the bin with the lower bound $v$). This way, the first tuples in the order are those values (lower bound of bins of values) of those situation features that intervention on them may have (potentially) the most effect on the value of the class situation feature.

\subsection{SEM Inference}\label{sec::csi}

Here we show how to infer the SEM of a given situation feature table in two steps:
\begin{itemize}
    \item The first step is \emph{causal structure discovery}, which involves discovering its causal structure of the situation feature table. This causal structure encodes the existence and the direction of the causal relationships among the situation features in the situation extraction plan of the given situation feature table.
    \item The second step is \emph{causal strength estimation}, which involves estimating a set of equations describing how each situation feature is influenced by its immediate causes. Using this information we can generate the SEM of the given situation feature table.
\end{itemize}
In the sequel, we describe these two steps.

\subsubsection{Causal Structure Discovery.}\label{sec::firstStep} The causal structure of the situation features in a given situation feature table can be determined by an expert who possesses the domain knowledge about the underlying process and the causal relationships between its features. But having access to such knowledge is quite rare. Hence, we support discovering the causal structure in a data-driven manner.

Several search algorithms have been proposed in the literature (e.g., \cite{chickering2002optimal,spirtes2000causation,OgarrioSR16}). The input of a search algorithm is observational data in the form of a situation feature table (and possibly knowledge) and its output is a graphical object that represents a set of causal structures that cannot be distinguished by the algorithm. One of these graphical objects is \emph{Partial Ancestral Graph (PAG)} introduced in \cite{zhang08}. 

A PAG is a graph whose vertex set is $\V = \SF$ but has different edge types, including $\rightarrow,\leftrightarrow,\cra ,\multimapdotboth$. Similar to $\twoheadrightarrow$, we use infix notation for $\rightarrow,\leftrightarrow,\cra ,\multimapdotboth$. Each edge type has a specific meaning. Let $\sitf_1, \sitf_2 \in \V$. The semantics of different edge types in a PAG are as follows:
\begin{itemize}
    \item $\sitf_1 \rightarrow \sitf_2$ indicates that $\sitf_1$ is a direct cause of $\sitf_2$.
    \item $\sitf_1 \leftrightarrow \sitf_2$ means that neither $\sitf_1$ nor $\sitf_2$ is an ancestor of the other one, even though they are probabilistically dependent (i.e., $\sitf_1$ and $\sitf_2$ are both caused by one or more hidden confounders).
    \item $\sitf_1 \cra \sitf_2$ means $\sitf_2$ is not a direct cause of $\sitf_1$.
   \item $\sitf_1 \multimapdotboth \sitf_2$ indicates that there is a relationship between $\sitf_1$ and $\sitf_2$, but nothing is known about its direction.
\end{itemize}

The formal definition of a PAG is as follows \cite{zhang08}: 
\begin{definition}[Partial Ancestral Graph (PAG)]
	A PAG is a tuple $(\V,\rightarrow,\leftrightarrow,\cra ,\multimapdotboth)$ in which $\V = \SF$ and $\rightarrow,\leftrightarrow,\cra,\multimapdotboth \subseteq \V \times \V$ such that $\rightarrow$, $\leftrightarrow$, $\cra$, and $\multimapdotboth$ are mutually disjoint.
\end{definition}

\begin{figure}[htb]
    \begin{subfigure}[c]{0.49\textwidth}
        \includegraphics[width=\textwidth]{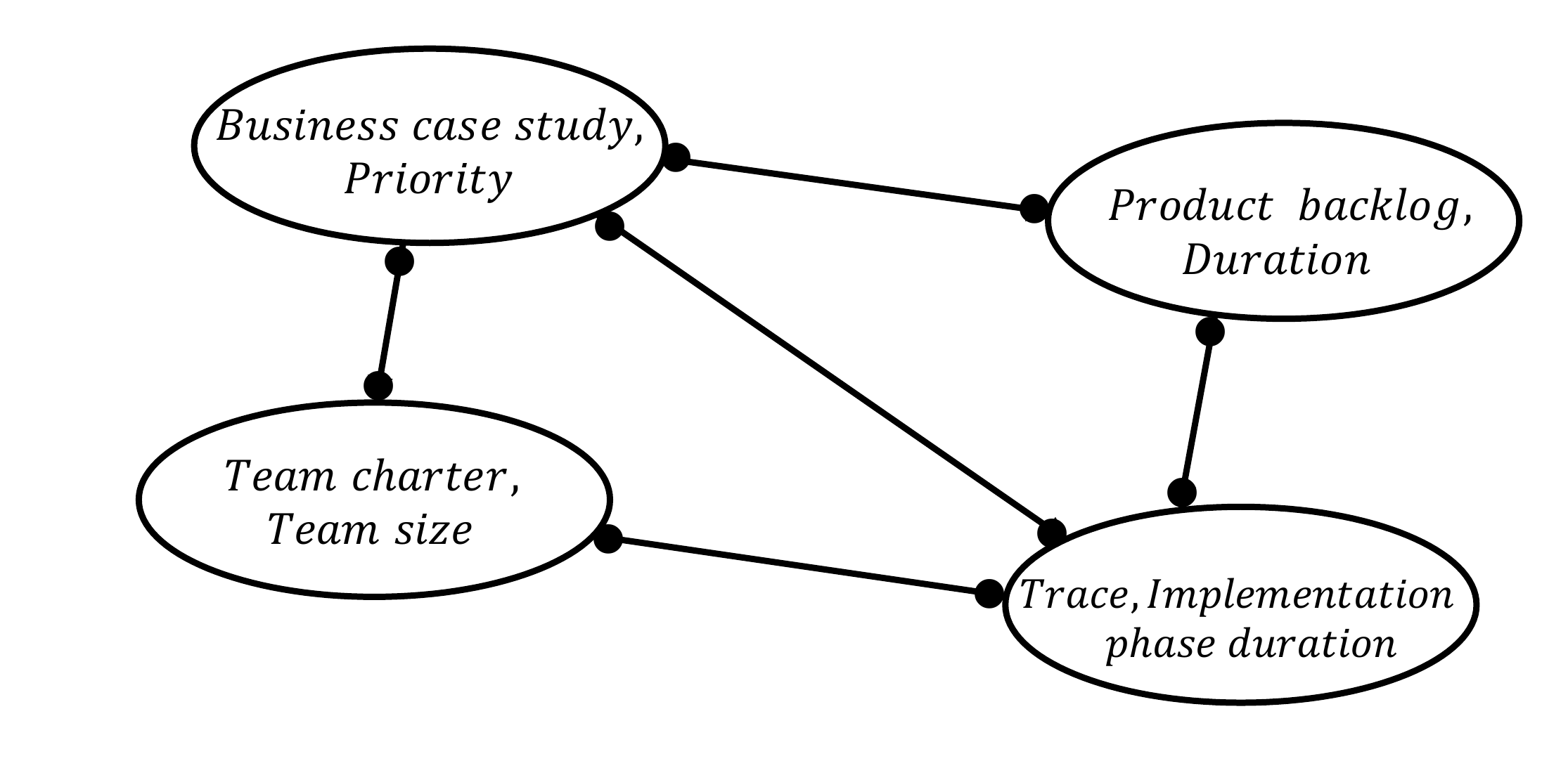}
        \subcaption{}
        \label{cg11}
    \end{subfigure}
    \begin{subfigure}[c]{0.49\textwidth}
        \includegraphics[width=\textwidth]{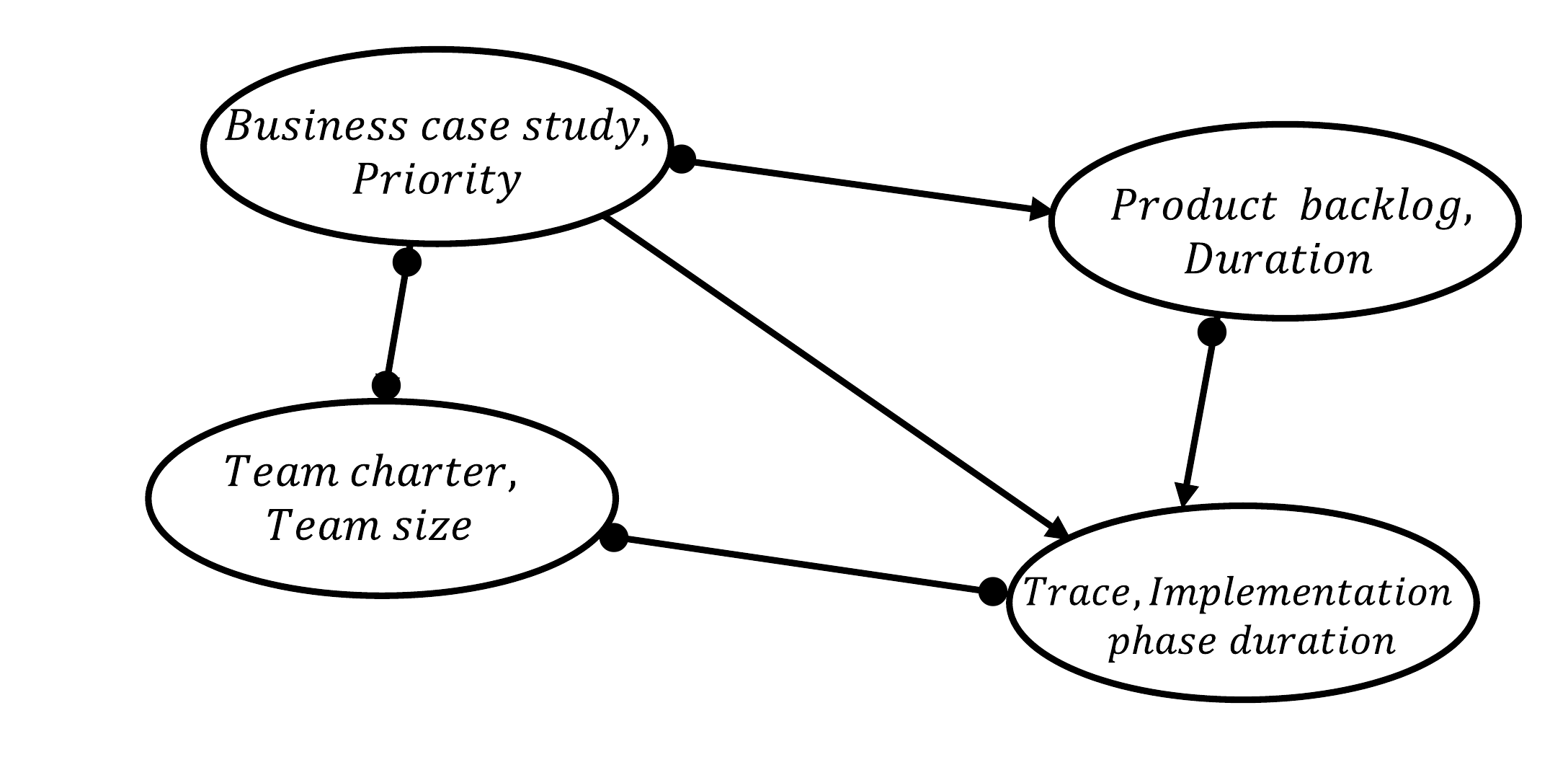}
        \subcaption{}
        \label{cg22}
    \end{subfigure}
    \caption{Two possible PAGs for the SEM mentioned in Example \ref{ex:sem}.}
    \label{fig::ex:pag}
\end{figure}

The discovered PAG by the search algorithm represents a class of causal structures that satisfies the conditional independence relationships discovered in the situation feature table and ideally, includes its true causal structure. 
\begin{example}\label{ex:pag}
Two possible PAGs for the SEM mentioned in Example \ref{ex:sem} are shown in Figure \ref{fig::ex:pag}.
\end{example}

Now, it is needed to modify the discovered PAG to a compatible causal structure. To transform the output PAG to a compatible causal structure, which represents the causal structure of the situation features in the situation feature table, domain knowledge of the process and common sense can be used. This information can be applied by directly modifying the discovered PAG or by adding them to the search algorithm, as an input, in the form of \emph{required directions} or \emph{forbidden directions} denoted as $D_{req}$ and $D_{frb}$, respectively. $D_{req}, D_{frb}\subseteq \V \times \V$ and $D_{req} \cap D_{frb} = \emptyset$. Required directions and forbidden directions influence the discovered PAG as follows:
\begin{itemize}
    \item If $(\sitf_1, \sitf_2) \in D_{req}$, then we have $\sitf_1 \to \sitf_2$ or $\sitf_1 \cra \sitf_2$ in the output PAG.
    \item If $(\sitf_1, \sitf_2) \in D_{frb}$, then in the discovered PAG it should not be the case that $\sitf_1 \to~\sitf_2$.
\end{itemize} 

We assume no hidden common confounder exists, so we expect that in the PAG, relation $\leftrightarrow$ be empty\footnote{If $\leftrightarrow \neq \emptyset$, the user can restart the procedure after adding some more situation features names to the situation feature table.}. We can define the compatibility of a causal structure with a PAG as follows:
\begin{definition}[Compatibility of a Causal Structure With a Given PAG]
	Given a PAG $(\V,\rightarrow,\leftrightarrow,\cra,\multimapdotboth)$ in which $\leftrightarrow = \emptyset$, we say a causal structure $(\U, \twoheadrightarrow)$ is compatible with the given PAG if $\V=\U$, $ ( \sitf_1 \rightarrow \sitf_2 \vee \sitf_1 \cra \sitf_2) \implies \sitf_1 \twoheadrightarrow \sitf_2$, and $\sitf_1 \multimapdotboth \sitf_2\implies (\sitf_1 \twoheadrightarrow \sitf_2 \oplus  \sitf_2 \twoheadrightarrow \sitf_1)$, where $\oplus$ is the XOR operation and  $ \sitf_1, \sitf_2 \in \V$.
\end{definition}
\begin{example}\label{ex:compatibleDAG}
The causal structure shown in  Figure \ref{fig:ex:causal Structure} is compatible with both PAGs demonstrated in Figure \ref{fig::ex:pag}.
\end{example}
\subsubsection{Causal Strength Estimation.}
The final step of discovering the causal model is estimating the strength of each direct causal effect using the observed data. Suppose $D$ is the causal structure of a situation feature table $T_{L,\SF,\csf}$. As $D$ is a directed acyclic graph, we can sort its nodes in a topological order $\gamma$. Now, we can statistically model each situation feature as a function of the noise terms of those situation features that appear earlier in the topological order $\gamma$ of $D$. In other words, $\sitf = f\big((N_{\sitf'})_{\sitf':\gamma(\sitf') \leq \gamma (\sitf)}\big)$ \cite{peters2017elements}. The set of these functions, for all $\sitf \in \SF$, is the SEM of $\SF$.

Finally, we want to answer questions about the effect of an intervention on any of the situation features on the class situation feature. We can do the intervention as described in Definition \ref{intervention}. The resulting SEM (after intervention) demonstrates the effect of the intervention on the situation features. 

Note that, in a given causal structure of a situation feature table $T_{L,\SF,\csf}$, there is no directed path between $\sitf \in \SF$ and $\csf$, they are independent and consequently, intervening on $\sitf$ by forcing $\sitf =c$ has no effect on $\csf$. 

\section{Experimental Results}\label{sec::er}
We have implemented the proposed approach as a plugin in ProM which is available in the nightly-build of ProM under the name \emph{Root-cause Analysis Using Structural Equation Model}. ProM is an open-source and extensible platform for process mining~\cite{verbeek2010prom}. The inputs of the implemented plugin are an event log, the Petri-net model of the process,
and, the conformance checking results of replaying the given event log on the given model. In the rest of this section, first, we mention some of the implementation details and design choices that we have made and then we present the results of applying the plugin on a synthetic and a real event log.

\subsection{Implementation Notes}
In the implemented plugin, we first enrich the event log by adding some attributes. Some of the features that can be extracted (besides the ones that have been explicitly mentioned in the event log) from the event log using the implemented plugin are as follows:
\begin{itemize}
    \item Time perspective: timestamp, activity duration, trace duration, trace delay, sub-model duration.
    \item Control-flow perspective: next activity, previous activity.
    \item Conformance perspective: deviation, number of log moves, number of model moves.
    \item Resource organization perspective: resource, role, group.
    \item Aggregated features:
    \begin{itemize}
        \item Process-level: the number of waiting customers, workload.
        \item Trace-level:average service time, average waiting time.
        \item Event-level: number of active events with a specific activity name, number of waiting events with a specific activity name.
        \item Resource-level: average service time, average waiting time
    \end{itemize}
\end{itemize}
As the second step, the user needs to specify $\csf$ and $\SF$. The user can specify $\SF$ by manually selecting the proper set of situation features or use the implemented situation feature recommendation method on a predefined set of situation features (for example all the situation features) to identify the relevant set of situation features to $\csf$. If $\SF$ includes an aggregated feature, we also compute the values of the aggregated feature
The third step involves extracting situation feature table from the event log. According to the selected $\SF$ and $\csf$ the proper situation subset of the event log is generated and the situation feature table is extracted. Then we infer the causal structure of the situation feature table. For this goal, we use the Greedy Fast Causal Inference (GFCI) algorithm \cite{OgarrioSR16} which is a hybrid search algorithm. The inputs of  GFCI algorithm are the situation feature table and possibly background knowledge. The output of GFCI algorithm is a PAG with the highest score on the input situation feature table. In \cite{OgarrioSR16}, it has been shown that under the large-sample limit, each edge in the PAG computed by GFCI is correct if some assumptions hold. Also, the authors of \cite{OgarrioSR16} using empirical results on simulated data have shown that GFCI has the highest accuracy among several other search algorithms. In the implemented plugin, we have used the Tetrad \cite{scheines1998tetrad} implementation of the GFCI algorithm. To use GFCI algorithm, we need to set several parameters. We have used the following settings for the parameters of the GFCI algorithm in the experiments: cutoff for p-values = 0.05, maximum path length = -1, maximum degree = -1, and penalty discount = 2.

In the implemented plugin, we have assumed linear dependencies among the situation features and additive noise when dealing with continuous data. In this case, given a SEM $\eq$ over $\SF$, we can encode $\eq$ as a weighted graph. This weighted graph is generated from the corresponding causal structure of $\eq$ by considering the coefficient of $\sitf_2$ in $\eq (\sitf_1)$ as the weight of the edge from $\sitf_2$ to $\sitf_1$. Using this graphical representation of a SEM, to estimate the magnitude of the effect of $\sitf$ on the $\csf$, we can simply sum the weights of all directed paths from $\sitf$ to $\csf$, where the weight of a path is equal to the multiplication of the weights of its edges. 

\subsection{Synthetic Event Log} \label{synth}
For the synthetic data, we use the IT company example in Section \ref{sec::ex}. The situation feature extraction plan is:
$$\{(Team\ size, G_3), (Duration, G_2),(Priority, G_1),(Implementation\ phase\ duration, \bot) \}.$$
where the class situation feature is $(Implementation\ phase\ duration, \bot)$.
We assume that the true causal structure of the data is as depicted in Figure \ref{cg3}.

\begin{figure}[t!]
	\includegraphics[width=120mm]{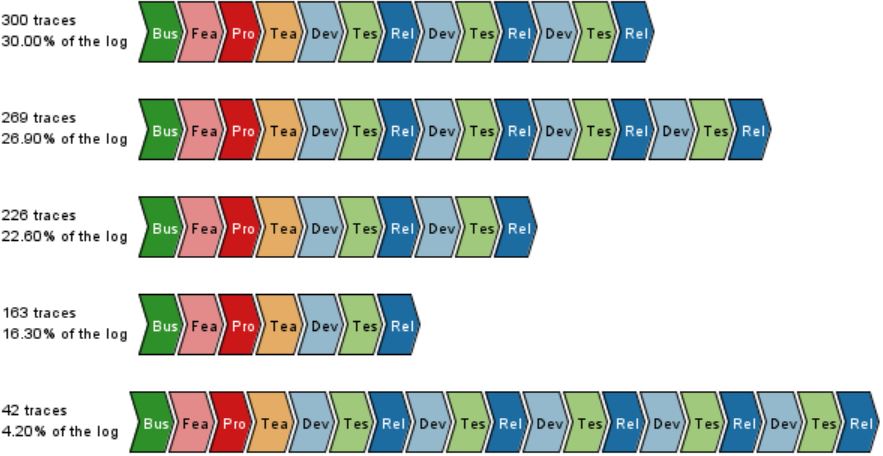}
	\caption{The trace variants in the synthetic event log.}\label{fig:logSummary}
\end{figure}

\begin{figure}[t!]
	\includegraphics[width=120mm]{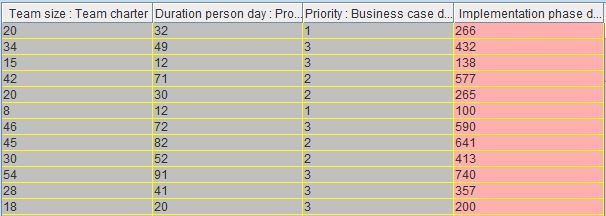}
	\caption{A snapshot of the situation feature table generated for the synthetic event log.}\label{fig:log1}
\end{figure}

\begin{figure}[htb]
    \begin{subfigure}[c]{0.49\textwidth}
        \includegraphics[width=\textwidth]{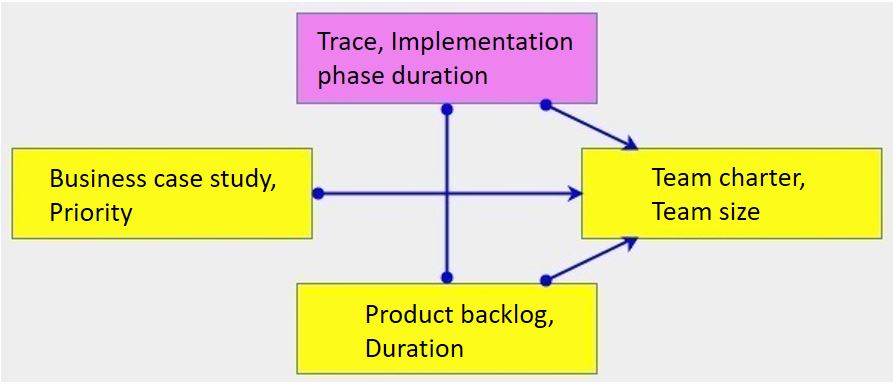}
        \subcaption{The PAG discovered by applying the implemented plugin on the situation feature table extracted using the mentioned situation feature extraction plane in this section.}
        \label{a}
    \end{subfigure}
    \begin{subfigure}[c]{0.49\textwidth}
        \includegraphics[width=\textwidth]{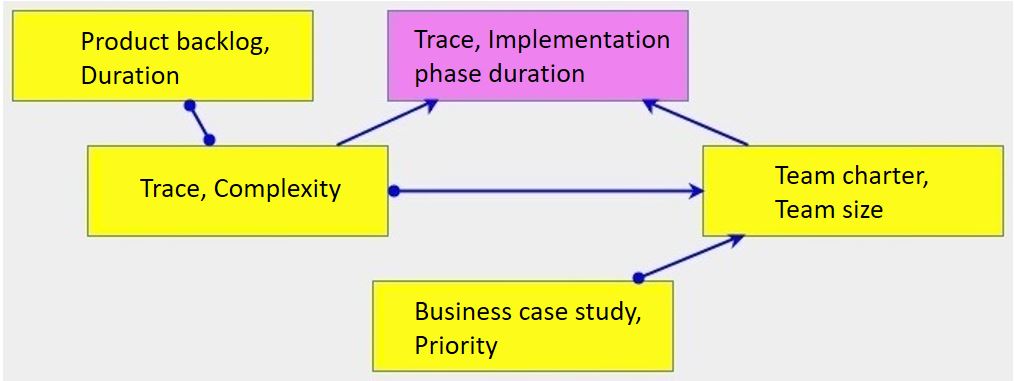}
        \subcaption{The discovered PAG after adding $(Complexity, \bot)$ to the situation feature extraction plan (as one of the descriptive situation features).}
        \label{b}
    \end{subfigure}
    \begin{subfigure}[c]{0.49\textwidth}
        \includegraphics[width=\textwidth]{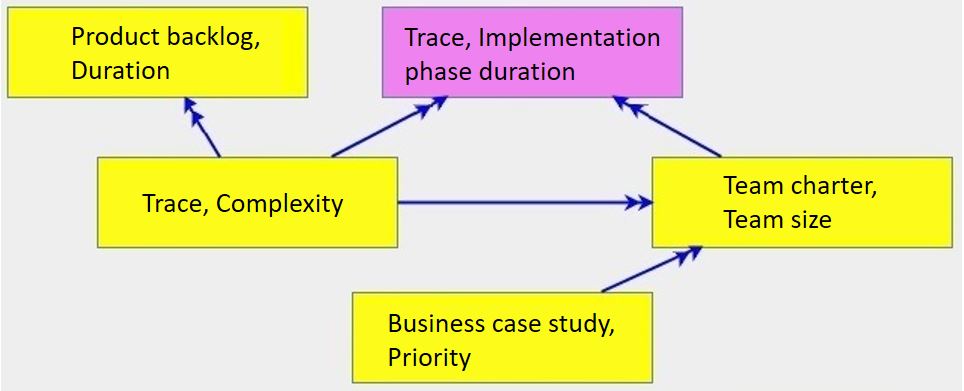}
        \subcaption{The causal structure which is obtained by modifying the PAG in \ref{b} based on common sense and domain knowledge.}
        \label{c}
    \end{subfigure}
    \begin{subfigure}[c]{0.49\textwidth}
        \includegraphics[width=\textwidth]{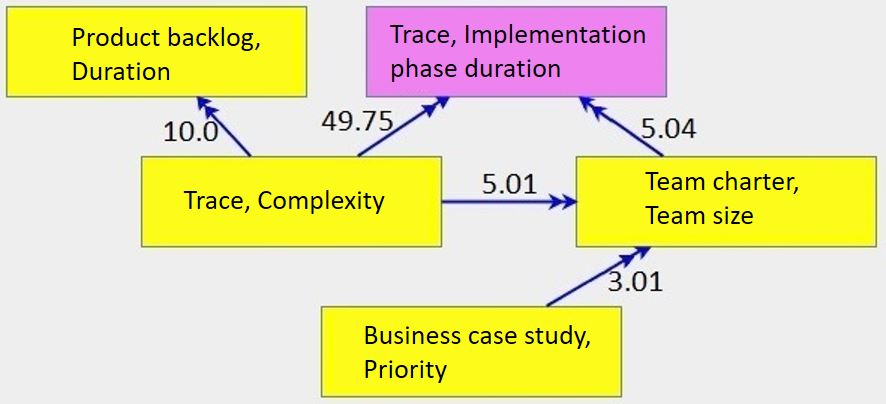}
        \subcaption{The inferred SEM by estimating the strength of the discovered causal relationships.}
        \label{d}
    \end{subfigure}
    \caption{The PAC, causal structure and the SEM discovered using implemented plugin for the synthetic event log.}
    \label{pic::exS}
\end{figure}

To generate an event log, we first created the Petri-net model of the process as shown in  \ref{pic::ex} using CPN 
Tools \cite{ratzer2003cpn}. Then, using the created model, we generated an event log with 1000 traces. We have enriched the event log by adding $Implementation\ phase\\ duration$ attribute to the traces. This attribute indicates the duration of the sub-model including ``development" and ``test" transitions in person-day. When generating the log, we have assumed that the true SEM of the process is as follows:
\begin{center}
	\resizebox{12cm}{!}{
		\begin{tabular}{ l l }
			$(Complexity, \bot)= N_{(Complexity, \bot)}$  & $N_{(Complexity, \bot)} \sim Uniform(1,10)$  \\
			
			$(Priority, G_1) = N_{(Priority, G_1)}$ &$N_{(priority, G_1)}\sim Uniform (1,3)$  \\
			
			$(Duration, G_2) =10 (Complexity, \bot) + N_{(Team\ size, G_3)}$ & $N_{(Duration, G_2)} \sim Uniform(-2,4)$  \\
			
			$(Team\ size, G_3) =5(Complexity, \bot) + 3(Priority, G_1) +N_{(Team\ size, G_3)}$ &   $N_{(Team\ size, G_3)} \sim Uniform(-1,2)$ \\
			
			$(Implementation\ phase\ duration, \bot) =50(Complexity, \bot) +  $  &$N_{(Implementation\ phase\ duration, \bot)} \sim Uniform(10,20)$  \\
			$5(Team\ size, G_3) +N_{(Implementation\ phase\ duration, \bot)}$& \\
	\end{tabular}}\label{eq}
\end{center} 
The summary of the generated event log and its trace variants (generated by ProM) are shown in Figure \ref{fig:logSummary}.
\paragraph{Generating situation feature table.}
We generate a situation feature table using the mentioned situation feature extraction plan. A snapshot of the generated situation feature using the implemented plugin is shown in Figure \ref{fig:log1}. In this figure, the class situation feature is colored in pink and the descriptive situation features are colored gray.
\paragraph{SEM inference.}
Applying the implemented plugin on the situation feature extracted from the event log, the PAG depicted in Figure \ref{a} was discovered. Even though the discovered PAG does a good job regarding discovering the potential causal relationship, it does not say much about the direction of them. Here the customer may guess that another influential attribute might exist that acts as a confounder. Considering $(Complexity, \bot)$ as another descriptive situation feature, then the discovered PAG by the implemented plugin would be as the one in Figure \ref{b}. This PAG is more accurate and includes the true causal structure of the situation feature table. We have assumed that the complexity of a project is a feature that is not recorded in the event log. The customer, may assume (based on domain knowledge) that the duration of ``product backlog" is longer in more complex projects and assign to the complexity of a project the floor of the value of $(Duration,G_2)$ divided by 10. Now, using domain knowledge and the chronological order of transitions, we can turn the discovered PAG into the causal structure depicted in Figure \ref{c}. After estimating the strength of the causal relationships, we obtain the SEM shown in Figure \ref{d}. 

\begin{figure}[t!]
	\includegraphics[width=120mm]{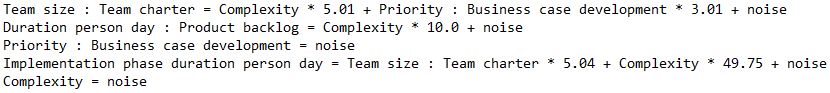}
	\caption{The discovered SEM from the situation feature table extracted from the synthetic event log after adding $(Complexity, \bot)$ to the situation feature extraction plan in the text format.}\label{semText}
\end{figure}

Moreover, we can have the inferred SEM in text format. In this case, the output would be as Shown in Figure \ref{semText}.

By comparing the estimated coefficients of situation features names in the output of the plugin (and equivalently the weights of the edges in Figure \ref{d}), and those in the equations of the true SEM of the data, we can see that the estimated and real strengths of causal relationships are quite close.

To investigate the effect of an intervention on any of the situation features on the class situation feature, we can find the equation capturing the effect of intervention by simply clicking on its corresponding node in the causal structure. For example, if we click on the corresponding node of $(Team\ size, G_3)$, we have $$(Implementation\ phase\ duration, \bot) = 75.0004 \times (Complexity, \bot) + noise.$$ This equation means that by enforcing the complexity of a project to be one unit more complex, then we expect that its implementation phase takes approximately 75 more person-days (assuming that the complexity of a project is actionable). As another example, equation $(Implementation\ phase\ duration, \bot)= 0.0 \times (Duration, G_2) $ shows the estimated effect of intervention on $(Duration, G_2)$. We can interpret this equation as ``intervention on $(Duration, G_2)$ has no effect on $(Implementation\ phase\ duration, \bot)$".

\subsection{Real Event Log}\label{real}
We have used the implemented plugin also on several real-life event logs. In this subsection we analyse \emph{receipt phase of an environmental permit application process (WABO) CoSeLoG project} \cite{buijs2014receipt} event log (receipt log for short). The receipt event log includes 1434 traces and 8577 events. 
The maximum duration of traces in the receipt event log is almost 276 days while and the average duration of traces is almost 3 minutes. We consider those traces that took longer than 1 percent of the maximum trace duration as delayed. Thus, the class situation feature is $(trace\ delay,\bot )$ and ``trace delay" is one of the trace-level attributes that has been used to enrich the receipt event log. The length of the time window in this experiment has been set to one day. 

\begin{figure}[t!]
	\includegraphics[width=80mm]{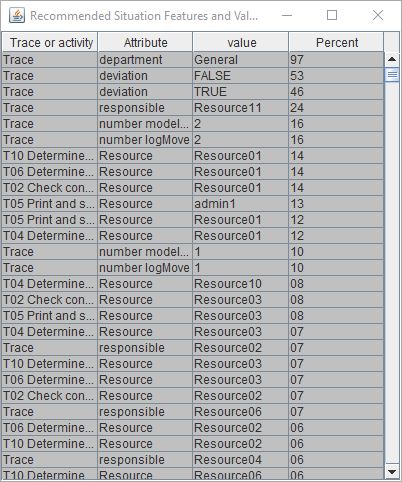}
	\caption{The first 27 recommended situation features and values by applying our new situation feature recommendation method.}\label{recomm}
\end{figure}

\paragraph{Situation feature extraction plan selection.} In this case study, we first use the situation feature recommendation in which we set the number of bins to 100. We use the set including features related to the resources of events, the duration of events, process workload, and some of the trace features (such as deviation, number of model moves, number of log moves, responsible, and department) as the initial set of features. In Figure \ref{recomm}, one can see the first 27 recommended situation features and values. The last column shows in which percent of the situations with the undesirable result, the situation feature has the mentioned value.

We set the $\alpha$ threshold to $0.05$\%. As a result, the situation feature extraction plan includes some of the resources for some of the events, responsible, deviation, number of log moves, number of model moves, and process workload. We have ignored some of the situation features recommended by the algorithm. For example, we have ignored $(department, \bot)$ as the value assigned to this situation feature in almost all of the traces is ``General", so it does not contain any information about the class situation feature. Also, we have ignored $(deviation, \bot)$ as each of its assigned values appears in almost the same portion of the situations with the undesirable result. We have ignored $(Number\ logMove, \bot)$ as it has the same value as $(Number\ modelMove, \bot)$ in all of the generated instances. For the sake of simplicity, we use the following abbreviation for activity names in the rest of this section.
\begin{itemize}
    \item ``T02" instead of ``T02 Check confirmation of receipt".
    \item ``T04" instead of ``T04 Determine confirmation of receipt",
    \item ``T05" instead of ``T05 Print and send confirmation of receipt",
    \item ``T06" instead of ``T06 Determine necessity of stop advice",
    \item ``T10" instead of ``T10 Determine necessity to stop indication",
\end{itemize}
The situation feature extraction plan in this example is as follows:
\begin{align*}
    \{(G_{T02},Resource),&(G_{T04},Resource),(G_{T05},Resource),(G_{T06},Resource),(G_{T10},Resource),\\&(\bot, Trace\ delay), (\bot, Process\ workload), (\bot, Responsible)\\
    & (\bot, Deviation), (\bot, Number\ logMove)\}
\end{align*}
where the class situation feature is $(\bot, Trace\ delay)$ and
\begin{equation*}
\begin{aligned}[c]
(G_{T02}\coloneqq group(actName, \{ \text{``T02"}\}) \\
(G_{T04}\coloneqq group(actName, \{ \text{``T04"}\})\\
(G_{T10}\coloneqq group(actName, \{ \text{``T10"}\}). \\
\end{aligned}
\ \ \ 
\begin{aligned}[c]
(G_{T05}\coloneqq group(actName, \{ \text{``T05"}\})\\
(G_{T06}\coloneqq group(actName, \{ \text{``T06"}\})
\end{aligned}
\end{equation*}

\begin{figure}[t!]
	\includegraphics[width=120mm]{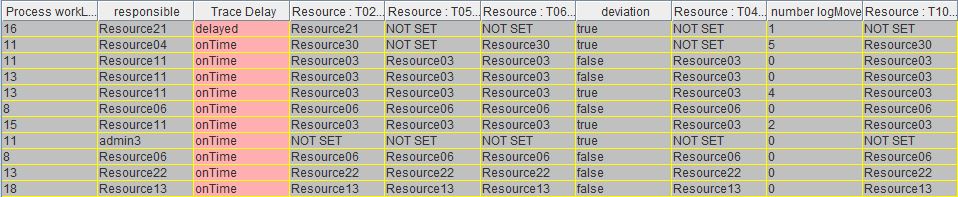}
	\caption{A snapshot of the situation feature table extracted from the receipt event log based on the selected situation feature extraction plan.}\label{tableReal}
\end{figure}

\paragraph{Generating situation feature table.}
Using the above situation feature extraction plan, we generate a situation feature table. A snapshot of the generated situation feature table is shown in Figure \ref{tableReal}.

\begin{figure}[t!]
	\includegraphics[width=120mm]{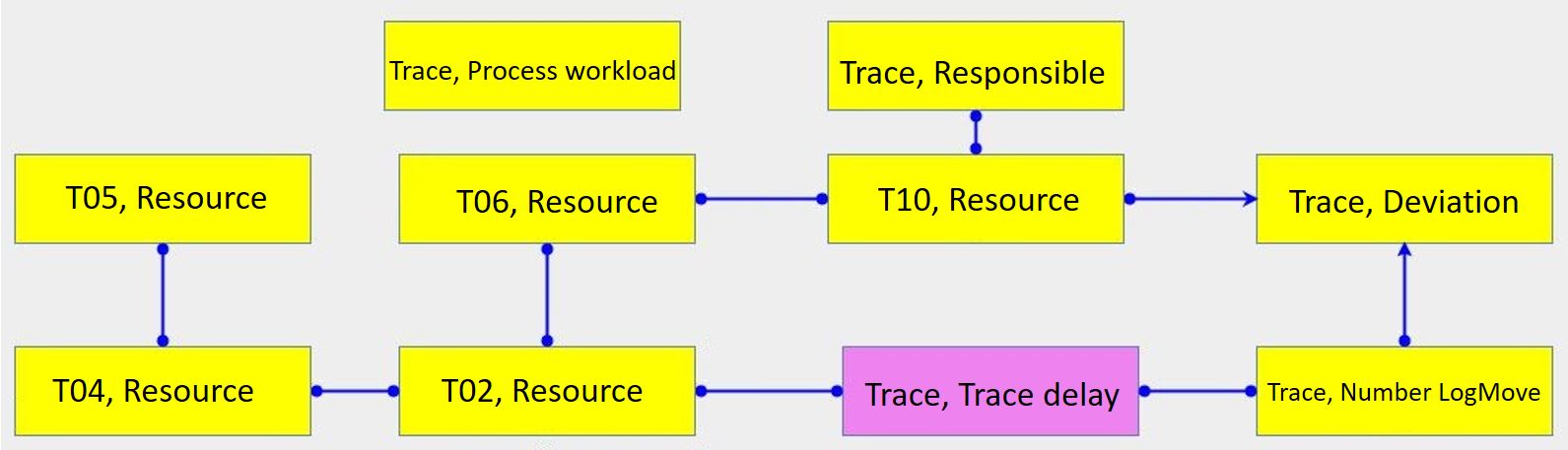}
	\caption{The PAG of the situation feature table extracted from the receipt event log, discovered by the implemented plugin. }\label{fig::pagRec}
\end{figure}

\begin{figure}[t!]
	\includegraphics[width=120mm]{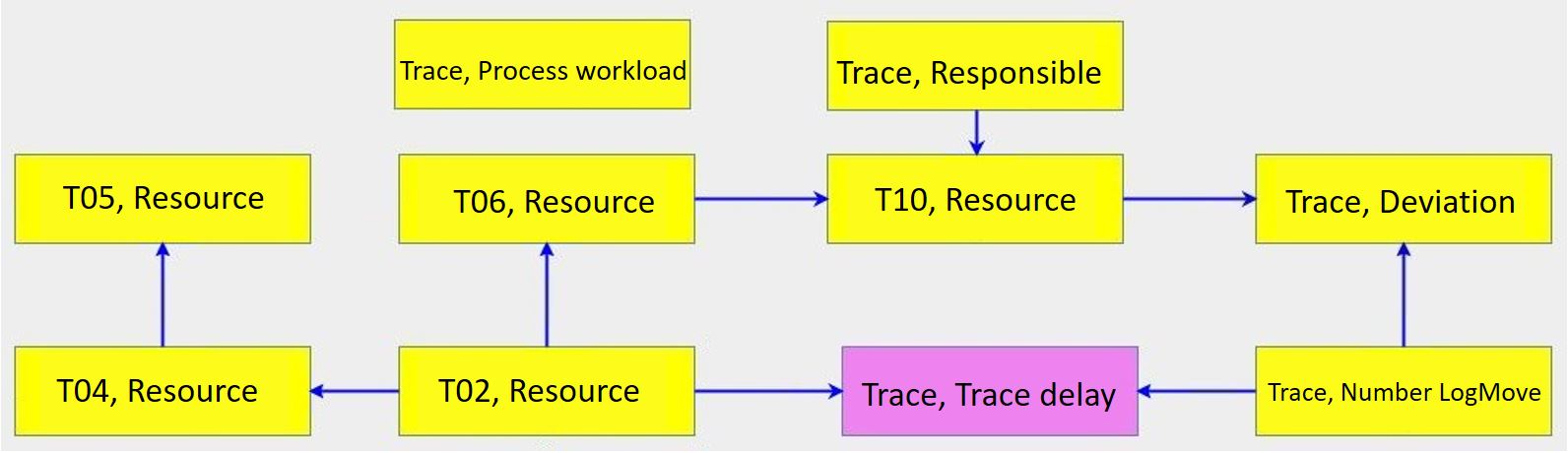}
	\caption{The causal structure obtained by modifying the PAG in Figure \ref{fig::pagRec} using common sense and domain knowledge.}\label{fig::semreal}
\end{figure}

\paragraph{SEM inference.}
The discovered PAG from the extracted situation feature is as shown in Figure \ref{fig::pagRec}. Using the temporal ordering of the activities (in this process, $T02$ happens before $T04$, $T04$ happens before $T05$, $T05$ happens before $T06$, $T06$ happens before $T10$ in all the traces) and common sense (the choice of the resource of an activity does not affect the choice of the resource of another activity that happened before) we are able to convert the PAG in Figure \ref{fig::pagRec} into the causal structure shown in Figure \ref{fig::semreal}.

According to this causal structure only $(G_{T02},Resource)$ and $(\bot, Number\ logMove)$ have a causal effect on $(\bot , Trace\ delay)$ and there is no causal relationship between other situation features and $(\bot , Trace\ delay)$. Some of the other causal relationships encoded in Figure \ref{fig::semreal} are as follows:
\begin{itemize}
    \item The choice of $(G_{T02},Resource)$ directly influences the choice of $(G_{T04},Resource)$ and $(G_{T06},Resource)$.
    \item The value of $(G_{T02},Resource)$ indirectly influences the value of $(G_{T05},Resource)$, $(G_{T10},Resource)$.
    \item $(\bot, responsible)$ and $(G_{T06},Resource)$ causally influence $(G_{T10},Resource)$.
    \item $(\bot, Number\ logMove)$ and $(G_{T10},Resource)$ directly influence $(\bot, Deviation)$.
    \item There is no causal relationships between any of the situation features and $(\bot,\\ Process \ workload)$.
\end{itemize}

\begin{figure}[t!]
	\includegraphics[width=80mm]{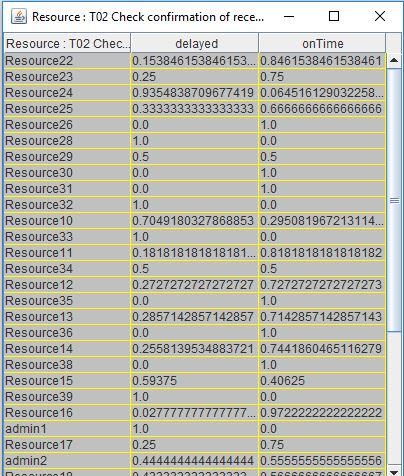}
	\caption{The predicted interventional distributions resulting from intervention on $(G_{T02},Resource)$ by enforcing different values.}\label{intDist}
\end{figure}

After applying the estimation step, we can see the interventional distributions of $(\bot ,Trace \ delay)$. We can see the effect of an intervention on any of the situation features by clicking on its corresponding node. For example, we can see that the probability of $(\bot , Trace \ delay) = delayed$ by forcing $(G_{T02},Resource)= Resource14$ is almost 0.256. The predicted interventional distributions resulting from enforcing $(G_{T02},Resource)$ to be assigned each of its possible values is shown in Figure \ref{intDist}.

It is worth noting that as the data in this case study is categorical, the inferred SEM includes several equations where the right hand side includes many terms. Each term shows the distribution of each one of the situation feature values condition on one of the values of one of its direct causes. So, it is not possible to completely present the inferred SEM like in the continuous case.
 
\section{Conclusion}\label{sec::conclution}
Distinguishing causal from mere correlational relationships among the process features is a vital task when investigating the root causes of performance and/or conformance problems in a company. The best way to identify the causal relationships is using randomized experiments. However, this requires implementing process changes to see their effect. As applying randomized experiments is usually quite expensive (if not impossible) in the processes, we propose a method for root cause analysis based on the theory of causality which uses a mixture of data analysis and domain knowledge. The stakeholders can use this framework to incorporate both domain knowledge and potential statistically supported causal effects to find the SEM of the features and indicators of the process. Moreover, this method helps stakeholders investigating the effect of an intervention on the process. This information can be used to design and order the re-engineering steps. 

The validity of a discovered structural equation model (and any other machine learning technique) is highly influenced by the set of features that have been used for data extraction and structural equation model discovery. However, the complex and dynamic inter-dependencies in processes makes the task of selecting the set of features with a potential causal effect on the observed problem in the process a challenging task. So, we have proposed a simple yet intuitive and effective feature recommendation method in this paper. The proposed method provides the user not just the set of features with the possible causal effect on the class situation feature but also those values of the features that potentially have the largest contribution to the observed problem in the process. It is worth noting that this was missing in the previous work on the causal structure model discovery of a process. 

As future work, we would like to learn more process- related features that go beyond individual cases. For example, bottlenecks are caused by competing cases or a shortage of resources. Also, notions such as blocking, batching, and overtaking are not captured well. We would also like to make the diagnostics more understandable. This requires mapping diagnoses related to features back onto the process model and even log. Finally, we would like to enhance simulation models with SEM-based rules.


\section*{Acknowledgement}
We thank the Alexander von Humboldt (AvH) Stiftung for supporting our research.

 \bibliographystyle{splncs04}
 \bibliography{biblio}
\end{document}